\documentclass[11pt]{article}

\usepackage[preprint]{acl}

\usepackage{times}
\usepackage{latexsym}
\usepackage[T1]{fontenc}
\usepackage[utf8]{inputenc}
\usepackage{microtype}
\usepackage{inconsolata}
\usepackage{graphicx}

\usepackage{booktabs}       
\usepackage{longtable}      
\usepackage{multirow}       
\usepackage{amsfonts}       
\usepackage{amsmath}        
\usepackage{nicefrac}       
\usepackage{xcolor}         
\usepackage{pifont}         
\newcommand{\cmark}{{\color{green!60!black}\ding{51}}}  
\newcommand{\xmark}{{\color{red!70!black}\ding{55}}}    
\newcommand{\pmark}{%
  \textcolor{orange}{%
    \ding{51}%
    \kern-0.65em%
    \raisebox{0.05em}{\ding{55}}%
  }%
}
\usepackage{caption}        
\usepackage{listings}       
\usepackage{upquote}        
\usepackage{tcolorbox}       
\tcbuselibrary{breakable, skins}
\newtcolorbox{taskbox}[1]{%
  colback=blue!3!white,
  colframe=blue!55!black,
  boxrule=0.4pt,
  arc=1mm,
  left=2mm, right=2mm, top=1mm, bottom=1mm,
  before skip=3pt, after skip=3pt,
  fonttitle=\bfseries\small,
  coltitle=white,
  title={#1},
  before upper={\raggedright\sloppy}
}
\usepackage{enumitem}       
\usepackage{tikz}           
\usetikzlibrary{shapes.geometric, arrows.meta, positioning, fit, backgrounds, calc}

\title{\raisebox{-0.3em}{\includegraphics[height=1.5em]{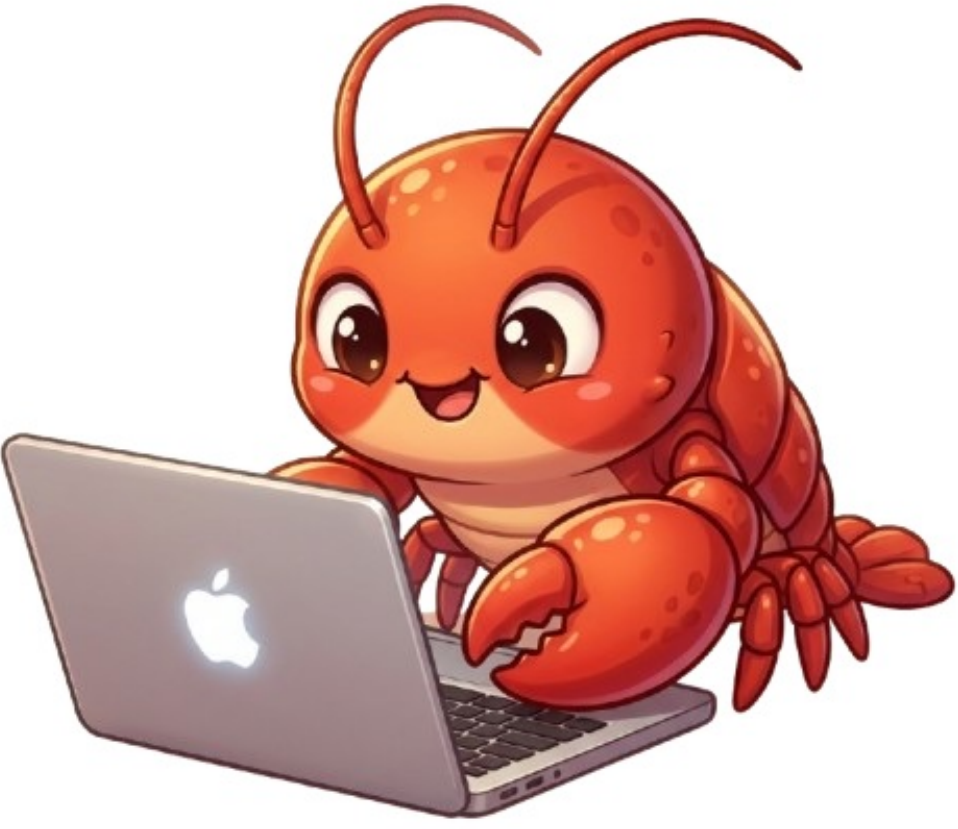}}~MacAgentBench: Benchmarking AI Agents on Real-World macOS Desktop}

\author{
  Yikun Fu$^{1,2}$\thanks{Equal contribution.}, Bowen Fu$^{3}$\footnotemark[1], Zhenyu Wu$^{1,2}$, Shuang Cheng$^{2,6}$, Xiaowei Sun$^{2,4}$, Bowen Yang$^{2,5}$, Zehao Li$^{2,5}$ \\  
  \bfseries Yibo Zhao$^{2,7}$, Zichen Ding$^{2}$, Zhoumianze Liu$^{2,4}$, Shijie Wang$^{2}$\footnotemark[2], Biqing Qi$^{2}$\thanks{Corresponding authors: \texttt{qibiqing@pjlab.org.cn}}, Bowen Zhou$^{2,8}$\footnotemark[2] \\[4pt]  
  {\normalfont\small
  $^{1}$Shanghai Jiao Tong University \quad
  $^{2}$Shanghai AI Laboratory \quad
  $^{3}$Xi'an Jiaotong University \quad
  $^{4}$Fudan University} \\
  {\normalfont\small
  $^{5}$University of Science and Technology of China \quad
  $^{6}$Zhejiang University \quad
  $^{7}$East China Normal University \quad
  $^{8}$Tsinghua University}
}

\begin{document}

\maketitle
\begin{abstract}
Computer use agents (CUAs) have advanced rapidly in desktop automation,
and a growing number of users deploy CUAs such as OpenClaw on Mac Mini
for always-on automation.
However, existing benchmarks, including those for macOS, evaluate
agents without framework augmentation and rely on binary evaluation.
As a result, they fail to capture both the framework
capabilities leveraged by modern CUAs and the partial progress
on long-horizon, multi-application tasks.
We present MacAgentBench, a comprehensive macOS agent benchmark
comprising 676 tasks across 25 applications, with nearly 60\% involving
both GUI and CLI interaction.
The benchmark adopts deterministic rule-based evaluation and introduces
fine-grained multi-checkpoint scoring with capability annotations for
multi-application tasks.
Experiments across three frameworks and 16 models show that the best
configuration, Claude Opus~4.6 on OpenClaw, attains 73.7\%
Pass@1, while this advantage is primarily driven by the skill library
rather than by framework design.
Fine-grained metrics further reveal that models with similar Pass@1
can differ substantially in sub-goal completion.
Our code and data are publicly available at \url{https://github.com/JetAstra/MacAgentBench}.
\end{abstract}

\section{Introduction}
\label{sec:intro}

\begin{figure*}[!htbp]
  \centering
  \includegraphics[width=\textwidth]{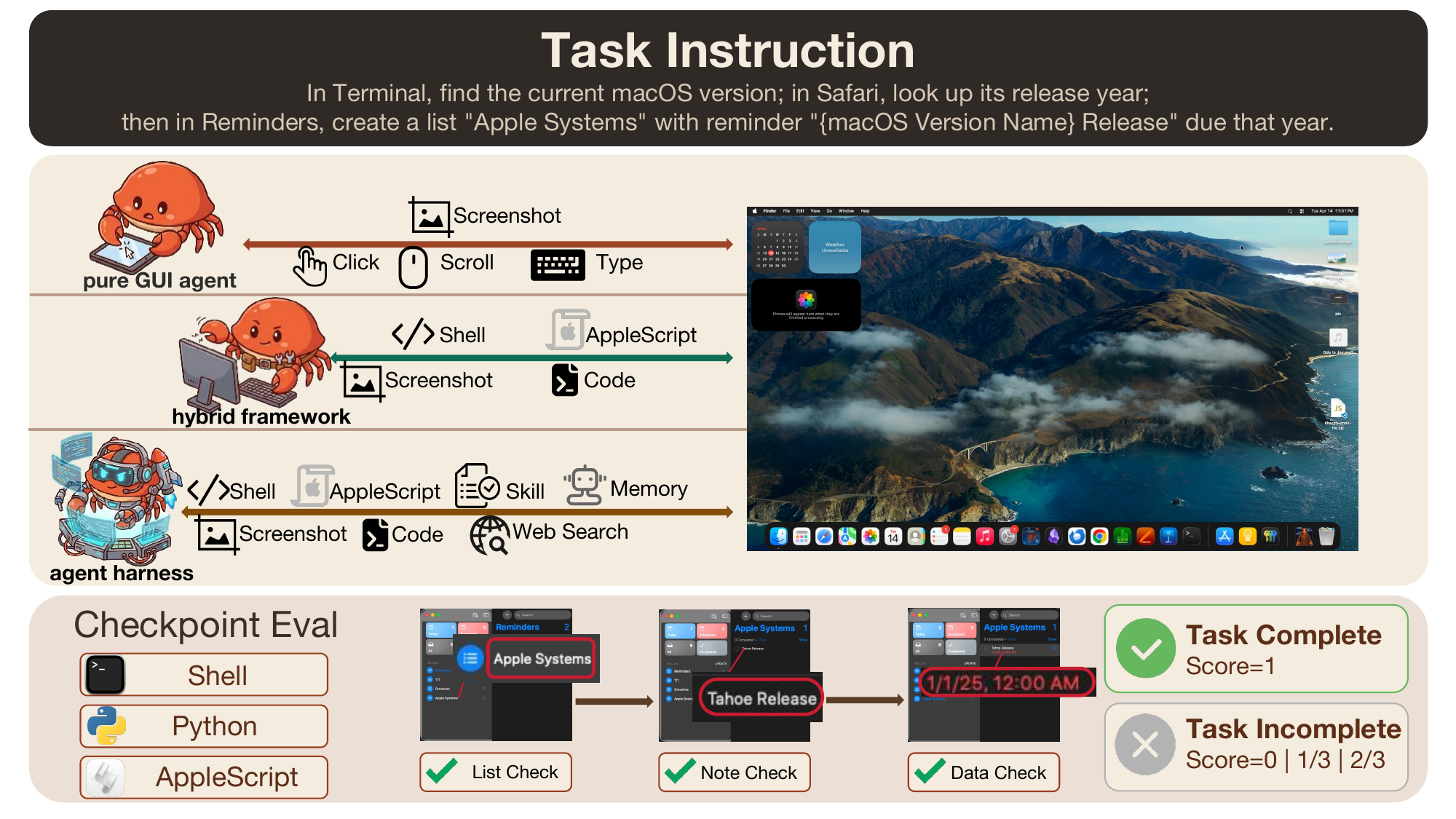}
  \caption{Overview of MacAgentBench. Three agent paradigms operate on a shared containerized macOS environment with progressively richer action spaces: a pure GUI agent, a hybrid framework, and an agent harness. Each task is scored by deterministic rule-based checkpoints, yielding partial credit on multi-application tasks.}
  \label{fig:overview}
\end{figure*}

Computer use agents (CUAs) that autonomously operate desktop environments
have emerged as a central focus of industry and academia.
Major model providers have released desktop control
capabilities~\citep{Claude3, Claude4, InternVL35,
bai2025qwen3vltechnicalreport, glm5team2026glm5vibecodingagentic,
UITARS, CogAgent},
and a growing ecosystem of CUA frameworks and models has
formed~\citep{ReAct, OSSymphony, AgentS, AgentS2, openclaw2025}.
Recent benchmarks such as OSWorld~\citeyearpar{OSWorld}
and WindowsAgentArena~\citeyearpar{WindowsArena}
systematically evaluate CUA on Linux and Windows,
driving standardization in the field.
Real-world deployment, however, is increasingly shifting to macOS,
a leading platform for software development, creative work, and
professional productivity.
An increasing number of users deploy CUAs such as OpenClaw on Mac Mini for always-on automation,
and integrate them into daily document, design, and engineering workflows.
macOS is uniquely suited for CUA deployment,
with a native layered automation stack: AppleScript for application scripting,
the Accessibility API for UI access, and a Unix command line.
This enables agents to choose the most efficient interaction method for each task.
Although prior macOS benchmarks~\citep{macOSWorld,macOSArena} have made initial explorations, they remain few, suffer from limited usability, 
and inherit the limitations of existing CUA benchmarks.

Specifically, existing CUA benchmarks
share two fundamental limitations.
First, they reduce CUA evaluation to a fixed screenshot--GUI-action loop.
Yet modern CUAs consist of a \emph{framework} and a \emph{model} working in concert,
in which the framework provides capabilities beyond GUI,
such as command-line access, scripting, and pre-built skills.
Second, they rely on binary pass/fail judgments.
For complex multi-application tasks, such coarse-grained evaluation fails to reveal how much progress an agent actually achieves.

To address this, we propose MacAgentBench, a comprehensive benchmark for evaluating CUAs on real-world macOS desktop tasks.
The fundamental design principle of MacAgentBench is \textbf{environment openness}: the benchmark imposes no restrictions on how agent frameworks interact with the environment, enabling fair comparison across diverse CUA paradigms, ranging from pure GUI agents to hybrid frameworks to agent harnesses, on the same set of tasks.
Built on a lightweight Docker containerized environment with quick startup and full task-level isolation, 
MacAgentBench comprises 676 tasks, nearly 60\% of which involve both GUI and CLI.
All evaluations employ deterministic rule-based scripts to ensure reproducibility.
Beyond binary pass/fail judgments, we further introduce fine-grained multi-checkpoint scoring for multi-application tasks, in which each checkpoint is annotated with one of five capability dimensions: Research, App State, Content, File Ops, and Sys Config.
This exposes partial progress and capability imbalance that binary metrics cannot capture.

We conduct extensive experiments across three frameworks and 16 models.
Our results reveal that framework design has a substantial impact on task completion:
the best configuration, Claude Opus~4.6 on OpenClaw, achieves 73.7\% Pass@1,
compared to 39.2\% for the same model without framework support.
To understand the source of this gain, we analyze the effect of pre-defined skills:
on tasks with skill coverage, OpenClaw substantially outperforms the baseline;
on tasks without skills, however, OpenClaw underperforms the baseline
for most models, revealing that its advantage is primarily skill-driven.
Through robustness analysis using Pass@1, Pass@4, and Pass$^4$ metrics,
we find that agents with similar capability ceilings
can differ substantially in consistency.
Furthermore, fine-grained multi-checkpoint evaluation reveals that
models with identical pass/fail rates can differ significantly
in sub-goal completion, exposing capability imbalances missed by binary evaluation.

Our main contributions are as follows:
\begin{itemize}
  \item We introduce MacAgentBench, the most comprehensive macOS agent benchmark to date,
    comprising 676 tasks with deterministic rule-based evaluation.
    Its open environment lets diverse CUA paradigms be evaluated
    without restriction, and nearly 60\% of tasks involve multiple interaction methods.
  \item Leveraging this open design, we conduct the first direct comparison on macOS
    across CUA paradigms spanning pure GUI agents, hybrid frameworks, and agent harnesses, 
    revealing the significant and independent impact of
    framework design and skill augmentation on agent performance.
  \item Since binary pass/fail evaluation cannot capture partial progress on long-horizon
    multi-application tasks, we introduce a multi-checkpoint evaluation mechanism
    that provides fine-grained reward signals for progress tracking and pinpointing
    the exact step at which an agent fails.
\end{itemize}

\section{Related Work}
\label{sec:related}

\paragraph{End-to-end CUA models.}
Advances in vision-language models (VLMs)~\cite{Claude4, bai2025qwen3vltechnicalreport, InternVL35} have spurred growing interest in CUAs. Early works such as CogAgent~\citeyearpar{CogAgent} and SeeClick~\citeyearpar{SeeClick} framed the problem as GUI grounding by eliciting VLM capabilities for interface understanding and interaction. Subsequent studies~\cite{Aguvis, OS-ATLAS, UGround, AgentTrek} explored diverse training paradigms to improve agents’ planning and decision-making. More recently, native agent models~\cite{UITARS, UI-TARS-2, InfiGUI-R1} have further advanced GUI perception and interaction through multi-stage training and reinforcement learning.

\paragraph{CUA frameworks.}
Despite progress, end-to-end CUA models still struggle with long-horizon tasks due to memory decay, limited generalization, and restricted action spaces. To mitigate these challenges, recent works propose framework-based agentic systems that separate high-level planning from low-level execution~\cite{AgentS, AgenticLybic, OSCopilot}. Some approaches enhance memory for long-horizon tasks via dedicated memory modules or context-control mechanisms~\cite{MGA, AgentProg}. Others improve open-environment generalization by integrating retrieval-augmented generation or searcher modules for dynamic external knowledge access~\cite{RAG-GUI, OSSymphony}. Yet others expand the action space through code agents, enabling hybrid GUI and CLI interaction for complex system operations~\cite{CoAct-1, UltraCUA}.

\paragraph{Benchmarks for CUAs.}
Existing benchmarks evaluate CUAs from different perspectives. ScreenSpot~\citeyearpar{SeeClick} and
ScreenSpot-Pro~\citeyearpar{ScreenSpot-Pro} focus on grounding, while others target desktop~\cite{OSWorld, WindowsArena}, mobile~\cite{AndroidWorld}, and web environments~\cite{WebArena, WebVoyager, Mind2Web}, primarily through GUI interactions. Beyond GUI tasks,
Terminal-Bench~\citeyearpar{Terminal-Bench} and SWE-Bench~\citeyearpar{SWE-bench} assess terminal-based capabilities, whereas ToolBench~\citeyearpar{ToolLLM} and $\tau$-bench~\citeyearpar{taubench} evaluate API and external tool usage. However, most existing
benchmarks, including those for macOS~\cite{macOSArena, macOSWorld}, vary the model under a fixed agent framework, despite real-world CUAs
consisting of a framework and a model working in concert. MacAgentBench fills this gap by evaluating multiple CUA frameworks on macOS.

\section{MacAgentBench}
\label{sec:benchmark}

\begin{table*}[!htbp]
  \caption{Comparison with existing agent benchmarks.
  \cmark\ = supported, \xmark\ = not supported, \pmark\ = partial.}
  \label{tab:task_compare}
  \centering
  \scalebox{0.88}{
  \begin{tabular}{llcccccc}
    \toprule
    Benchmark & Platform & Tasks & Task Type & Multi-App & Rule Eval & Checkpoints & Task Var. \\
    \midrule
    Mind2Web~\citeyearpar{Mind2Web} & Web & 2,350 & GUI & \xmark & \cmark & \xmark & \xmark \\
    WebArena~\citeyearpar{WebArena} & Web & 812 & GUI & \cmark & \pmark & \xmark & \cmark \\
    VisualWebArena~\citeyearpar{VisualWebArena} & Web & 910 & GUI & \cmark & \pmark & \xmark & \cmark \\
    WebVoyager~\citeyearpar{WebVoyager} & Web & 643 & GUI        & \xmark & \xmark & \xmark & \xmark \\
    Online-Mind2Web~\citeyearpar{Online-Mind2Web} & Web & 300 & GUI        & \xmark & \xmark & \pmark & \xmark \\
    EconWebArena~\citeyearpar{DBLP:journals/corr/abs-2506-08136} & Web & 360 & GUI        & \xmark & \cmark & \xmark & \cmark \\
    ClawBench~\citeyearpar{clawbench} & Web & 153 & GUI        & \xmark & \xmark & \xmark & \xmark \\
    OSWorld~\citeyearpar{OSWorld} & Linux  & 369    & GUI, CLI    & \cmark & \cmark & \pmark & \xmark \\
    TheAgentCompany~\citeyearpar{TheAgentCompany} & Linux & 175 & GUI, CLI    & \cmark & \pmark & \cmark & \xmark \\
    SWE-bench~\citeyearpar{SWE-bench} & Linux & 2,294 & CLI        & \xmark & \cmark & \xmark & \xmark \\
    Terminal-Bench~\citeyearpar{Terminal-Bench} & Linux & 89 & CLI        & \pmark & \cmark & \xmark & \xmark \\
    Claw-Eval~\citeyearpar{claweval} & Linux & 300 & CLI        & \cmark & \pmark & \cmark & \xmark \\
    GAIA~\citeyearpar{GAIA}    & Linux     & 466    & CLI        & \cmark & \cmark & \xmark & \xmark \\
    PinchBench~\citeyearpar{PinchBench} & Linux & 53 & CLI        & \cmark & \pmark & \cmark & \xmark \\
    WindowsAgentArena~\citeyearpar{WindowsArena} & Windows & 154 & GUI, CLI & \xmark & \cmark & \xmark & \xmark \\
    AndroidWorld~\citeyearpar{AndroidWorld} & Android & 116 & GUI & \cmark & \cmark & \pmark & \cmark \\
    ToolBench~\citeyearpar{ToolLLM} & API & 1,100 & API        & \pmark & \xmark & \xmark & \xmark \\
    $\tau$-bench~\citeyearpar{taubench} & API & 165 & API        & \xmark & \cmark & \xmark & \xmark \\
    macOSWorld~\citeyearpar{macOSWorld} & macOS & 202 & GUI, CLI & \cmark & \cmark & \xmark & \pmark \\
    macOSArena~\citeyearpar{macOSArena} & macOS & 63 & GUI, CLI & \cmark & \cmark & \xmark & \xmark \\
    \midrule
    \textbf{MacAgentBench} & \textbf{macOS} & \textbf{676} & \textbf{GUI, CLI} & \cmark & \cmark & \cmark & \cmark \\
    \bottomrule
  \end{tabular}
  }
\end{table*}

MacAgentBench provides a complete macOS desktop as a testbed for agent evaluation. The underlying infrastructure is a macOS virtual machine running inside a Docker container. The benchmark imposes no restrictions on how agent frameworks access the environment, supporting both in-container and remote deployment. It comprises 676 tasks across 25 applications, nearly 60\% of which involve both GUI and CLI. All evaluations employ deterministic rule-based methods, and multi-application tasks support fine-grained multi-checkpoint scoring. Table~\ref{tab:task_compare} compares MacAgentBench with existing benchmarks.

\subsection{Infrastructure}
\label{sec:env_infra}

MacAgentBench virtualizes a macOS environment using a Docker-QEMU stack. A key optimization leverages the copy-on-write mechanism of QEMU: all containers share a single base image, and each instance records only runtime writes in a lightweight differential layer. This achieves a 30\,s container startup time and a 1\,GB per-instance disk overhead, enabling parallel evaluation across multiple instances on a single server.

In contrast, macOSWorld uses AWS EC2 Mac bare-metal instances. Each task reset requires a 15\,min restore from an AMI snapshot on cloud-hosted dedicated hardware, hindering local reproduction. macOSArena also adopts a Docker-QEMU stack but requires a full copy of the 40\,GB disk image per container, yielding a 5\,min startup time and disk overhead that scales linearly with the number of parallel instances. A detailed comparison is provided in Appendix~\ref{app:env_infra} (Table~\ref{tab:env_compare}).

\subsection{Agent-environment interaction}
\label{sec:agent_env}

Building on this infrastructure, MacAgentBench imposes no restrictions on
agent-environment interaction.
Agent frameworks can either run inside the container or operate externally.
The observation space, action space, and interaction loop are determined
entirely by the agent framework.

We identify two representative deployment modes.
(1) \textbf{In-container deployment}: frameworks such as OpenClaw are pre-installed inside the container as a native macOS application. Once launched with a task instruction, the application orchestrates the full AI loop internally, assembling context, invoking the model, and executing tool calls, with direct access to screen capture, file system, AppleScript, and shell commands. Through this orchestration, the model can leverage any macOS functionality.
(2) \textbf{External deployment}: frameworks run outside the container and interact with the environment remotely. They retrieve observations needed by the model, such as screenshots and command output, from the macOS instance, and translate predicted actions into executable operations on the environment.

This open design distinguishes MacAgentBench from prior benchmarks that
focus exclusively on GUI-based agent interaction.

\begin{figure*}[ht]
\vspace{-6pt}
    \centering
    \begin{minipage}{0.48\textwidth}
        \centering
        \captionof{table}{Statistics of MacAgentBench.}
        \scalebox{0.85}{
        \begin{tabular}{l@{\hskip 8em}c}
            \toprule
            \textbf{Category} & \textbf{Tasks} \\
            \midrule
            \textbf{Total} & \textbf{676 (100\%)} \\
            - Productivity  & 224 (33.1\%) \\
            - System        & 44 (6.5\%) \\
            - Internet      & 108 (16.0\%) \\
            - Development   & 44 (6.5\%) \\
            - Multimedia    & 116 (17.2\%) \\
            - Multi-App     & 140 (20.7\%) \\
            \midrule
            \textbf{Interaction} & \\
            - GUI           & 124 (18.3\%) \\
            - CLI           & 148 (21.9\%) \\
            - GUI+CLI       & 404 (59.8\%) \\
            \midrule
            \textbf{Instructions} & \\
            Avg. length     & 28.1 words \\
            \bottomrule
        \end{tabular}
        }
        \label{tab:task_distribution}
    \end{minipage}
    \hspace{5mm}
    \begin{minipage}{0.46\textwidth}
        \centering
        \includegraphics[width=0.995\textwidth]{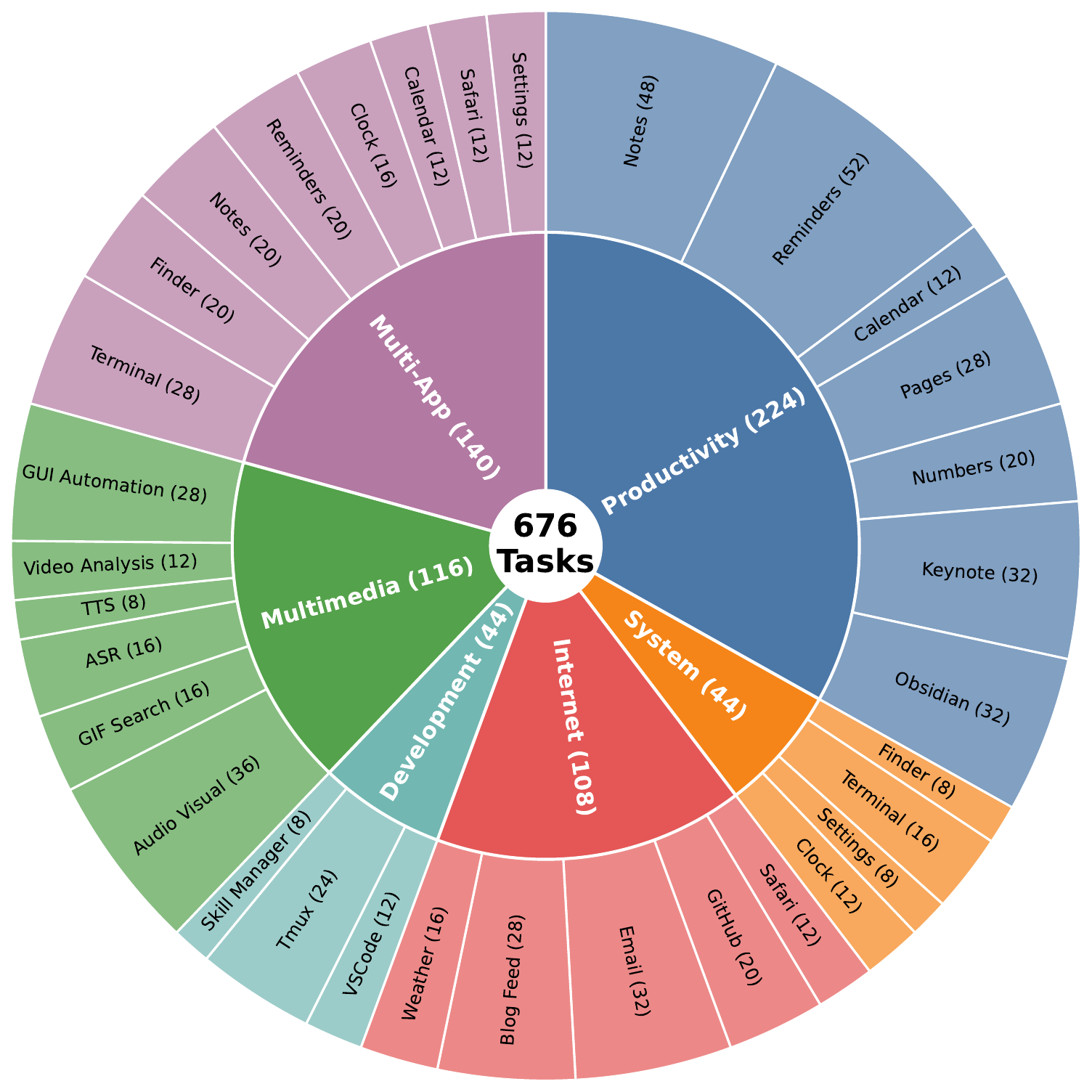}
        \caption{Distribution of MacAgentBench.}
        \label{fig:task_taxonomy}
    \end{minipage}
    \vspace{-9.5pt}
\end{figure*}

\subsection{Task construction}
\label{sec:task_construction}

 MacAgentBench comprises 676 tasks spanning 25 applications. The task design follows two core principles. First, it covers diverse interaction methods including CLI, GUI, and combinations of both, reflecting the variety of real-world macOS workflows. Second, it spans multiple complexity levels, from single-application operations to cross-application workflows.

\paragraph{Seed tasks.}
\label{sec:seed_tasks}

The seed tasks in MacAgentBench are drawn from two sources. The first set comprises 63 tasks adapted from macOSArena~\citeyearpar{macOSArena}. Because the macOS version has been upgraded, AppleScript interfaces have changed and application UI hierarchies have been restructured, rendering the original evaluation scripts incompatible. We verified each task individually and rewrote the evaluation scripts to address both types of changes. The second set comprises 110 newly designed tasks spanning two categories. One targets the iWork suite of Pages, Numbers, and Keynote, which are among the most widely used productivity applications on macOS. The other targets frequent desktop operations such as organizing email and checking weather forecasts. After cross-validation, we excluded 4 tasks whose evaluators were unreliable due to underlying macOS bugs. The full distribution of the final 169 seed tasks is shown in Table~\ref{tab:task_distribution}.

\paragraph{Task specification.}
Each task is specified as a triplet:
a \emph{natural-language instruction} that describes the goal,
an \emph{environment setup script} that configures the macOS instance
to the required initial state,
and an \emph{evaluation script} that inspects the final environment state
and returns a pass/fail result.
For multi-application tasks, the evaluation script also produces
a fine-grained multi-checkpoint score (see Section~\ref{sec:evaluation}).
Appendix~\ref{app:annotation} provides details of the annotation
and cross-validation process.

\paragraph{Template-based task expansion.}
\label{sec:task_expansion}

Each seed task is expanded into 4 variants to introduce
variation in both parameters and phrasing,
yielding the full set of $169 \times 4 = 676$ tasks.
The expansion follows a two-stage process.
The first stage, \emph{parameter substitution}, alters concrete values in the task (such as file names, dates, and content strings) and
updates the setup and evaluation scripts accordingly.
The second stage, \emph{instruction rewriting}, uses an LLM to rephrase
the task instruction while preserving the original task objective,
producing semantically equivalent but lexically distinct variants.
Each variant therefore differs from the seed task
in both task parameters and instruction phrasing.

\subsection{Task statistics}
\label{sec:task_stats}

As shown in Figure~\ref{fig:task_taxonomy} and Table~\ref{tab:task_distribution},
MacAgentBench spans six categories across 25 applications:
Productivity, with apps such as Notes, Pages, and Keynote;
System, with Terminal and Settings;
Internet, with GitHub and Email;
Development, with VS Code and Tmux;
Multimedia, with ASR and TTS;
and Multi-App, with cross-application workflows.
This breadth makes MacAgentBench the largest macOS benchmark to date,
with 404 tasks (59.8\%) involving both GUI and CLI.
\subsection{Evaluation protocol}
\label{sec:evaluation}

MacAgentBench employs a fully deterministic, rule-based evaluation protocol,
avoiding the reproducibility issues of LLM-as-judge methods.
Each evaluator consists of two components:
a \emph{getter} function that extracts the relevant state from the environment,
and a \emph{metric} function that compares the extracted result against
the expected value and returns a pass/fail result.
A single task may invoke multiple getters to verify different aspects
of task completion.
In total, 156 unique getter functions cover all 676 tasks
and fall into three categories:

\begin{itemize}
  \item \textbf{Shell commands} (88 functions) read file contents, database state, and system configuration via utilities such as \texttt{cat} and \texttt{grep}.
  \item \textbf{AppleScript} (48 functions) queries the state of macOS-native applications (e.g., Calendar events, Keynote properties) via \texttt{osascript}.
  \item \textbf{Python scripts} (20 functions) handle more complex verification logic such as file content comparison and audio attribute checking.
\end{itemize}

All evaluators are deterministic: the same environment state
produces the same evaluation result.

\paragraph{Fine-grained multi-checkpoint evaluation.}
For multi-application tasks, a single pass/fail result cannot reveal
where the agent fails in a multi-step workflow.
MacAgentBench therefore defines fine-grained \emph{checkpoints}
for the 140 multi-application tasks listed in Table~\ref{tab:task_distribution}.
Each checkpoint corresponds to a key sub-goal of the task
and is evaluated by its own getter and metric, following the same evaluator architecture.
The number of checkpoints per task averages 4.1, ranging from 2 to 7.
The final score is the fraction of checkpoints passed, in $[0, 1]$;
a task is marked as passed only when all checkpoints succeed.
For example, the task ``create a calendar and add three events'' has four
checkpoints: whether the calendar exists, and whether each of the three events
has been added.
This enables fine-grained progress tracking: a score of $0.75$ indicates
that the agent completed three of the four sub-goals.

Each checkpoint is further annotated with a \emph{capability} tag
from one of five dimensions:
\emph{research} for external knowledge queries,
\emph{app\_state} for application state verification,
\emph{content\_match} for content correctness checks,
\emph{file\_ops} for file system operations,
and \emph{system\_config} for system setting modifications.
This annotation enables analysis of model strengths and weaknesses
across distinct capability dimensions.

\subsection{Evaluation pipeline}
\label{sec:eval_pipeline}

The evaluation of each task proceeds through five stages.
\textbf{(1)~Environment startup}: a fresh Docker container is launched for the task;
the virtual machine boots from a copy-on-write layer on the shared base image
and becomes ready within 30\,s.
\textbf{(2)~Task initialization}: a task-specific setup script is executed
to configure the required initial state, for example by creating files,
launching applications, or populating database entries.
\textbf{(3)~Agent execution}: the natural-language task instruction is delivered
to the agent, which then operates autonomously in the environment until it signals
completion or reaches the maximum step limit.
\textbf{(4)~Evaluation}: a deterministic, rule-based evaluator,
written as a shell command, AppleScript, or Python script, inspects
the final environment state and returns a pass/fail result.
\textbf{(5)~Reset}: the container is destroyed, discarding the copy-on-write layer;
the next task starts from a clean base image, ensuring full inter-task isolation.

\section{Experiments}
\label{sec:experiments}

\begin{table*}[!htbp]
  \caption{Main results on MacAgentBench.}
  \label{tab:main_results}
  \centering              
  \footnotesize
  \setlength{\tabcolsep}{6pt}      
  \renewcommand{\arraystretch}{0.90}
  \begin{tabular}{clcccccc}
    \toprule
    Framework & Model & Pass@1 (\%) & Pass@4 (\%) & Pass$^4$ (\%) & Steps & Tokens (In/Out) & Cost(\$) \\
    \midrule
    \multirow{16}{*}{\hfil /\hfil}
  & Claude Opus 4.6      & 39.2 & 48.5 & 28.4 & 27.1 & 177K / 4.3K & 0.98 \\                                  
  & GPT-5.4              & 58.4 & 79.9 & 36.7 & 13.2 & 127K / 3.0K & 0.36 \\                                  
  & Gemini 3.1 Pro       & 34.2 & 47.9 & 21.9 & 26.5 & 127K / 19.7K & 0.49 \\ 
  & Qwen3VL-235B-A22B    & 21.6 & 37.3 &  9.5 & 23.2 & 257K / 4.4K & 0.08 \\
  & Qwen3VL-32B          & 21.3 & 36.7 & 10.1 & 32.7 & 385K / 10.4K & 0.04 \\                                
  & Qwen3VL-8B           & 14.5 & 29.0 &  4.1 & 35.2 & 468K / 35.9K & 0.10 \\                                 
  & InternVL3.5-14B         &  6.4 & 14.2 &  0.6 & 32.9 & 119K / 5.2K & -- \\   
  & InternVL3.5-8B          &  4.7 & 10.7 &  1.2 & 32.1 & 115K / 4.7K & -- \\ 
  & UI-TARS-72B-DPO      & 13.2 & 23.1 &  5.3 & 29.2 & 432K / 2.6K & -- \\
  & UI-TARS-1.5-7B       &  9.8 & 18.3 &  3.6 & 42.2 & 651K / 4.1K & 0.07 \\
  & ScaleCUA-32B         & 10.5 & 20.7 &  2.4 & 33.5 & 135K / 4.1K & -- \\
  & ScaleCUA-7B          &  6.7 & 13.0 &  1.2 & 30.3 & 121K / 4.3K & -- \\
  & GUI-Owl-1.5-32B      & 17.2 & 34.3 &  6.5 & 27.6 & 323K / 5.3K & -- \\
  & GUI-Owl-1.5-8B       & 10.2 & 20.7 &  2.4 & 24.8 & 284K / 9.3K & -- \\
  & OpenCUA-32B          & 18.8 & 34.9 &  7.7 & 20.1 & 176K / 8.3K & -- \\
  & OpenCUA-7B           & 15.1 & 29.0 &  5.9 & 24.1 & 214K / 10.1K & -- \\         
    \midrule
    \multirow{8}{*}{AgentS3}
    & Claude Opus 4.6      & 66.9 & 82.2 & 50.3 & 10.0 & 262K / 4.4K  & 1.30  \\
    & GPT-5.4              & 58.9 & 79.3 & 36.7 & 10.2 & 263K / 3.0K  & 0.66  \\
    & Gemini 3.1 Pro       & 54.3 & 70.4 & 37.9 & 8.1 &  136K / 8.7K & 0.36 \\
    & Qwen3VL-235B-A22B    & 46.7 & 68.6 & 27.2 & 13.3 & 379K / 16.8K & 0.14  \\
    & Qwen3VL-32B          & 37.4 & 58.6 & 17.2 & 13.4 & 379K / 20.5K & 0.05 \\
    & Qwen3VL-8B           & 28.6 & 45.0 & 13.0 & 13.7 & 426K / 29.3K & 0.05 \\
    & InternVL3.5-14B         & 13.5 & 30.8 &  3.6 & 13.6 & 515K / 5.1K & -- \\
    & InternVL3.5-8B          &  9.6 & 20.1 &  1.8 & 21.6 & 1075K / 12.3K & -- \\
    \midrule
  \multirow{8}{*}{OpenClaw}
  & Claude Opus 4.6      & 73.7 & 85.2 & 58.6 & 8.6 & 263K / 1.2K & 1.35 \\
  & GPT-5.4              & 60.7 & 82.8 & 38.5 & 6.5 & 153K / 1.5K & 0.40 \\                                  
  & Gemini 3.1 Pro       & 63.3 & 82.8 & 40.2 & 6.6 & 122K / 2.0K & 0.27 \\      
  & Qwen3VL-235B-A22B    & 46.0 & 63.9 & 28.4 & 8.4 & 211K / 5.7K & 0.07 \\   
  & Qwen3VL-32B          & 33.7 & 50.9 & 14.2 & 8.3 & 209K / 7.6K & 0.02 \\                                  
  & Qwen3VL-8B           & 26.9 & 45.0 & 13.6 & 13.2 & 455K / 12.7K & 0.07 \\                                
  & InternVL3.5-14B         &  0.0 &  0.0 &  0.0 & 1.0 & 8K / 0.8K & -- \\                                      
  & InternVL3.5-8B          &  0.0 &  0.0 &  0.0 & 1.0 & 8K / 0.2K & -- \\
    \bottomrule
  \end{tabular}
\end{table*}

\subsection{Experimental setup}
\label{sec:exp_setup}

We evaluate three agent frameworks. \textbf{/} represents a baseline pure GUI agent that uses only screenshots and mouse/keyboard actions; \textbf{AgentS3}~\citep{AgentS3} is a multi-agent collaboration architecture; and \textbf{OpenClaw}~\citep{openclaw2025} is an in-container agent harness with access to shell, AppleScript, and built-in skills. 
We evaluate 16 models, organized into general VLMs and native GUI agent models; the full list with citations and per-framework coverage appears in Appendix~\ref{app:models} (Table~\ref{tab:main_results}). General VLMs are evaluated on all three frameworks, while native GUI agent models are evaluated only on the baseline framework, as they cannot reliably produce the action formats required by AgentS3 and OpenClaw. Each task is capped at 50 interaction steps, and experiments cover the full 676-task set.

\paragraph{Evaluation metrics.}
We report the following metrics throughout the paper.
\textbf{Pass@1} is the fraction of tasks solved on a single attempt.
\textbf{Pass@4} is the fraction of tasks solved on at least one of four attempts.
\textbf{Pass$^4$} is the fraction of tasks solved on all four attempts.
\textbf{Score} is the fraction of checkpoints passed on multi-application tasks,
further decomposed along five capability dimensions:
Research, App State, Content, File Ops, and Sys Config.
\textbf{Avg.\ Steps} is the average number of interaction steps per task,
where a lower value indicates more efficient execution.
\textbf{Tokens (In/Out)} are the average input and output token counts per task.
\textbf{Avg.\ Cost} is the average API cost (USD) per task.

\subsection{Main results and analysis}
\label{sec:main_results}

\paragraph{Framework comparison.}
Keeping the model constant, AgentS3 delivers consistent improvements over the baseline across all models.
On Claude Opus~4.6, for example, Pass@1 increases from 39.2\% (baseline) to 66.9\% (AgentS3).
OpenClaw also outperforms the baseline on most models, with Pass@1 reaching 73.7\% on Claude Opus~4.6.
However, the gains from OpenClaw are less consistent than those from AgentS3.
On smaller models, the improvement from OpenClaw falls below that from AgentS3:
on Qwen3VL-32B, AgentS3 outperforms OpenClaw by 3.7 points.
On InternVL models, the additional context introduced by OpenClaw
interferes with these models, causing a sharp performance drop;
they fail to produce the tool-calling format OpenClaw requires,
rendering the framework unusable.

\paragraph{Model comparison.}
Across all frameworks, proprietary models substantially outperform open-source ones.
Claude Opus~4.6 on OpenClaw achieves the highest Pass@1 at 73.7\%.
On the baseline framework, GPT-5.4 achieves the highest Pass@1 at 58.4\%,
surpassing Claude Opus~4.6 at 39.2\%;
yet with framework support, Claude overtakes GPT-5.4 on both AgentS3 and OpenClaw,
suggesting that Claude benefits more from framework augmentation.

\paragraph{Robustness analysis.}
Substantial gaps exist between Pass@1, Pass@4,
and Pass$^4$, indicating that agents can solve many tasks
but fail to do so consistently across attempts.
This highlights robustness as a critical dimension for evaluating agent capability
beyond single-trial success rates.
Claude and GPT-5.4 on OpenClaw provide a revealing comparison:
both achieve close Pass@4, 85.2\% and 82.8\% respectively,
suggesting similar capability ceilings;
however, their Pass@1 differs by 13.0 points and Pass$^4$ by 20.1 points,
revealing that although the two models can solve a similar set of tasks,
Claude completes them far more consistently.

\subsection{Effect of skill augmentation}
\label{sec:skill_analysis}

\begin{table}[!ht]
  \caption{Pass@1 (\%) on tasks with and without skill coverage across frameworks.
  Subscripts on AgentS3 and \texttt{/} report $\Delta$ vs.\ OpenClaw.}
  \label{tab:skill_effect}
  \centering
  \small
  \resizebox{\columnwidth}{!}{%
  \begin{tabular}{llccc}
    \toprule
    Subset & Model & OpenClaw & AgentS3 & / \\
    \midrule
  \multirow{8}{*}{w/ Skill}
& Claude Opus 4.6          &  89.4 & $ 84.3_{\,-5.1}$ & $ 55.9_{\,-33.5}$ \\
& GPT-5.4                  &  77.9 & $ 72.6_{\,-5.3}$ & $ 67.6_{\,-10.4}$ \\
& Gemini 3.1 Pro           &  80.6 & $ 70.2_{\,-10.4}$ & $ 50.5_{\,-30.1}$ \\
& Qwen3VL-235B-A22B        &  64.6 & $61.4_{\,-3.2}$ & $28.2_{\,-36.4}$ \\       
& Qwen3VL-32B              &  52.9 & $ 47.1_{\,-5.9}$ & $ 28.7_{\,-24.2}$ \\
& Qwen3VL-8B               &  43.6 & $ 39.1_{\,-4.5}$ & $ 21.3_{\,-22.3}$ \\
& InternVL3.5-14B          &   0.0 & $ 19.9_{\,+19.9}$ & $  9.3_{\,+9.3}$ \\
& InternVL3.5-8B        &   0.0 & $ 14.6_{\,+14.6}$ & $  7.4_{\,+7.4}$ \\
  \midrule
  \multirow{8}{*}{w/o Skill}
& Claude Opus 4.6          &  45.6 & $ 48.8_{\,+3.1}$ & $ 16.2_{\,-29.4}$ \\
& GPT-5.4                  &  33.8 & $ 41.9_{\,+8.1}$ & $ 45.6_{\,+11.9}$ \\
& Gemini 3.1 Pro           &  43.1 & $ 37.5_{\,-5.6}$ & $ 13.8_{\,-29.4}$ \\
& Qwen3VL-235B-A22B        &  22.5 & $30.6_{\,+8.1}$ & $19.4_{\,-3.1}$ \\ 
& Qwen3VL-32B              &  13.1 & $ 30.6_{\,+17.5}$ & $ 15.6_{\,+2.5}$ \\
& Qwen3VL-8B               &   8.8 & $ 18.1_{\,+9.4}$ & $  9.4_{\,+0.6}$ \\
& InternVL3.5-14B             &   0.0 & $  8.8_{\,+8.8}$ & $  4.4_{\,+4.4}$ \\
& InternVL3.5-8B              &   0.0 & $  5.6_{\,+5.6}$ & $  2.5_{\,+2.5}$ \\
    \bottomrule
  \end{tabular}}
\end{table}

\begin{table*}[!htbp]
  \caption{Fine-grained evaluation results on multi-application tasks.}
  \label{tab:fine_eval}
  \centering
  \footnotesize
  \setlength{\tabcolsep}{6pt}      
  \renewcommand{\arraystretch}{0.90}
  \begin{tabular}{clccccccc}
    \toprule
    Framework & Model & Pass@1 (\%) & Score & Research & App State & Content & File Ops & Sys Config \\
    \midrule
    \multirow{16}{*}{\hfil /\hfil}
  & Claude Opus 4.6      & 20.7 & 31.4 & 24.6 & 31.5 & 15.8 & 81.4 &  0.0 \\                                                       
  & GPT-5.4              & 48.6 & 64.1 & 55.0 & 70.7 & 56.1 & 97.7 & 90.0 \\                                                       
  & Gemini 3.1 Pro       & 13.6 & 26.7 & 25.1 & 33.2 & 12.2 & 62.8 &  0.0 \\     
  & Qwen3VL-235B-A22B    &  6.4 & 19.1 &  7.9 & 25.5 &  9.4 & 53.5 & 25.0 \\
  & Qwen3VL-32B          &  7.9 & 18.7 & 12.6 & 21.7 &  8.6 & 34.9 & 55.0 \\                                                       
  & Qwen3VL-8B           &  2.1 &  9.0 &  1.6 & 12.5 &  3.6 & 25.6 & 20.0 \\                                                       
  & InternVL3.5-14B       &  0.7 &  9.3 &  0.5 & 16.8 &  4.3 & 18.6 & 10.0 \\                                                       
  & InternVL3.5-8B    &  0.0 &  5.4 &  0.0 &  8.2 &  2.9 & 18.6 &  0.0 \\                                                       
  & UI-TARS-72B-DPO      &  1.4 & 12.9 &  3.1 & 15.2 &  8.6 & 25.6 & 35.0 \\                                                       
  & UI-TARS-1.5-7B       &  1.4 &  8.9 &  1.0 & 14.7 &  5.0 & 18.6 &  5.0 \\ 
  & ScaleCUA-32B         &  0.7 & 12.3 &  5.2 & 22.8 &  6.5 & 23.3 &  5.0 \\                                 
  & ScaleCUA-7B          &  0.0 &  7.8 &  3.1 & 12.0 &  2.9 & 23.3 &  0.0 \\ 
  & GUI-Owl-1.5-32B      &  7.9 & 19.6 & 13.6 & 19.0 &  8.6 & 55.8 & 45.0 \\      
  & GUI-Owl-1.5-8B           &   3.6 &  15.8 &   6.3 &  23.9 &   3.6 &  48.8 &   5.0 \\
  & OpenCUA-32B              &   7.9 &  21.9 &  11.0 &  25.5 &  12.9 &  53.5 &  50.0 \\
  & OpenCUA-7B               &   1.4 &  13.6 &   4.2 &  20.1 &  10.1 &  41.9 &   5.0 \\                                                         
    \midrule
    \multirow{8}{*}{AgentS3}
  & Claude Opus 4.6      & 40.7 & 53.1 & 35.3 & 47.6 & 30.4 & 100.0 & 66.7 \\                                                      
  & GPT-5.4              & 41.4 & 59.3 & 52.4 & 64.7 & 45.3 & 95.3 & 85.0 \\                                                       
  & Gemini 3.1 Pro       & 30.7 & 50.6 & 50.3 & 55.4 & 31.7 & 90.7 &   35.0 \\                                                       
  & Qwen3VL-235B-A22B    &  25.7 & 39.2 & 28.3 & 40.8 &  26.6 & 72.1 & 70.0 \\                                                       
  & Qwen3VL-32B          & 19.3 & 33.0 & 25.1 & 28.8 & 20.1 & 65.1 & 65.0 \\    
  & Qwen3VL-8B           &  12.1 & 25.5 &   15.2 &   27.2 &   12.2 &   58.1 &   40.0 \\                                                       
  & InternVL3.5-14B         &  1.4 & 10.0 &  3.1 & 15.2 &  4.3 & 30.2 &  5.0 \\                                                     
  & InternVL3.5-8B          &  0.7 &  7.6 &  1.6 & 13.6 &  3.6 & 16.3 &  0.0 \\
    \midrule
    \multirow{8}{*}{OpenClaw}
  & Claude Opus 4.6      & 63.6 & 75.0 & 75.4 & 77.2 & 80.6 & 90.7 & 60.0 \\                                                       
  & GPT-5.4              & 45.0 & 56.3 & 54.5 & 54.3 & 50.4 & 90.7 & 30.0 \\                                                       
  & Gemini 3.1 Pro       & 40.0 & 51.9 & 52.4 & 51.1 & 57.6 & 93.0 & 30.0 \\    
  & Qwen3VL-235B-A22B    & 22.9 & 42.6 & 39.3 & 46.7 & 43.2 & 72.1 & 30.0 \\
  & Qwen3VL-32B          &  5.7 & 21.8 & 14.1 & 26.6 & 23.0 & 48.8 & 10.0 \\                                                       
  & Qwen3VL-8B           &  2.9 & 19.4 & 12.0 & 23.4 & 10.1 & 58.1 & 10.0 \\                                                       
  & InternVL3.5-14B         &  0.0 &  4.9 &  0.0 &  8.7 &  2.9 & 16.3 &  0.0 \\                                                       
  & InternVL3.5-8B          &  0.0 &  4.9 &  0.0 &  8.7 &  2.9 & 16.3 &  0.0 \\ 
    \bottomrule
  \end{tabular}
\end{table*}

To understand whether the advantage of OpenClaw comes from its skill library
or its framework design, we identify two subsets of tasks: those covered by pre-defined skills and those without (Table~\ref{tab:skill_effect}).
Tasks with ambiguous skill coverage are excluded from this analysis.
On skill-covered tasks, OpenClaw leads by a large margin:
with Claude, it reaches 89.4\% vs.\ 84.3\% for AgentS3
and 55.9\% for the baseline, confirming that skills provide substantial gains.
On tasks without skill coverage, however, this advantage largely disappears.
For 5 out of 8 models, OpenClaw even underperforms the baseline,
with the baseline column showing positive $\Delta$ values.
This indicates that the overall lead in the main results
is primarily driven by its skill library rather than its framework design alone.
In contrast, AgentS3, which does not use skills,
consistently outperforms the baseline on both task subsets.
For example, on tasks without skill coverage, Claude on AgentS3 reaches 48.8\% vs.\ 16.2\% on the baseline,
demonstrating that multi-agent collaboration alone
provides stable performance gains.

\subsection{Fine-grained evaluation analysis}
\label{sec:fine_eval_analysis}

Table~\ref{tab:fine_eval} reports the fine-grained evaluation on 140
multi-application tasks.
The Score column consistently exceeds Pass@1,
revealing that agents often make partial progress on sub-goals even when the overall task fails.
For example, GPT-5.4 on the baseline achieves 48.6\% Pass@1 but 64.1\% Score,
indicating that many failed tasks reflect partial success rather than total failure.
This distinction is further illustrated by
UI-TARS-72B-DPO and UI-TARS-1.5-7B on the baseline:
both achieve the same 1.4\% Pass@1,
yet their Scores differ, at 12.9\% and 8.9\% respectively.
Pass/fail evaluation treats them as equally poor,
while checkpoint-based scoring reveals that the larger model
completes meaningfully more sub-goals.
Some checkpoints verify that the agent does not corrupt
the pre-existing environment state, such as preserving untouched files
or avoiding incorrect labels. That is why even completely unusable
combinations such as InternVL3.5 on OpenClaw receive a small non-zero Score
despite achieving 0\% Pass@1.

Across capability dimensions, File Ops Scores are consistently the highest,
indicating that file system operations are the easiest for agents.
In contrast, Research Scores are generally the lowest,
suggesting that querying external knowledge remains the primary bottleneck.

The capability profiles also differ across frameworks.
Claude on OpenClaw achieves a relatively balanced profile, 
with Scores ranging from 60.0\% on Sys Config to 90.7\% on File Ops.
In contrast, GPT-5.4 on the baseline shows a highly uneven profile:
97.7\% on File Ops and 90.0\% on Sys Config,
but only 55.0\% on Research and 56.1\% on Content,
revealing that strong performance on certain dimensions
can mask weaknesses in others that only fine-grained evaluation can expose.

\section{Conclusion}
\label{sec:conclusion}

We presented MacAgentBench, a comprehensive benchmark for evaluating
CUAs on real-world macOS desktop tasks.
Built on a lightweight containerized environment with task-level isolation,
MacAgentBench provides deterministic, rule-based evaluations and supports
fine-grained multi-checkpoint evaluation with capability annotations on multi-application tasks,
exposing imbalanced capability profiles
invisible to pass/fail metrics.
Framework design substantially affects performance,
but the lead of OpenClaw is driven primarily by its skill library rather than its framework design.
Moreover, the gap between Pass@1 and Pass$^4$ highlights robustness
as a critical, under-explored dimension.

\section*{Limitations}

The macOS system runs in a QEMU virtual machine without Apple hardware
GPU support. The resulting differences from native macOS are
predominantly rendering effects rather than functional differences in
application behavior (e.g., GPU-accelerated rounded corners and view
transitions). During task construction we compared each task against
macOS running on Apple hardware and excluded any task whose setup,
expected behavior, or evaluation outcome diverged between the two
environments. Because task success is defined over file-system,
AppleScript, and application-state outcomes rather than pixel-level
visuals, the residual rendering differences do not affect the success
criteria of any included task.
The benchmark is tied to macOS Tahoe~26.
Since application behaviors, AppleScript interfaces, and UI layouts
may change across macOS versions,
adapting MacAgentBench to a new version would require
reviewing and updating the setup and evaluation scripts.
MacAgentBench currently covers 676 tasks.
Although this is the largest macOS agent benchmark to date, it could be expanded to cover more applications and scenarios.

\paragraph{Potential risks.}
CUAs capable of GUI and CLI interaction can, in principle,
be misused to automate sensitive operations such as unauthorized file access
or credential harvesting if deployed on user systems without proper safeguards.
MacAgentBench substantially mitigates such risks by design.
First, all tasks execute within an isolated Docker-QEMU container with
copy-on-write disk layers discarded after each task,
ensuring no persistent state and no access to real user data;
the trajectories we release are therefore recordings of agent behavior
on synthetic, sandboxed task instances rather than real user activity.
Second, the task environment itself contains no personally identifying
information: as described in Appendix~\ref{app:license},
all third-party content was de-identified prior to inclusion, with real
names, email addresses, phone numbers, and file paths replaced by
fabricated placeholders.
We release the complete benchmark artifacts for all 676 tasks, including task instructions, setup scripts, deterministic evaluation scripts, checkpoint definitions, metadata, and environment configuration. We also release gold reference solutions for the subset of tasks whose canonical solutions can be expressed as CLI or AppleScript commands.
In addition, we release
agent execution trajectories (screenshots, actions, and checkpoint-level evaluation outcomes) for all 676 tasks.
These artifacts are intended for reproducibility, evaluator
validation, and failure-mode analysis (e.g., Appendix~\ref{app:failures}),
and are distributed under a research-use license. 
While the released trajectories could in principle be repurposed for
benchmark-specific tuning or overfitting, they are limited to benign,
sandboxed productivity tasks and contain no real user data, credentials,
or access tokens, thereby reducing privacy leakage and direct-abuse risks.
Third, we recommend that any CUAs evaluated against
MacAgentBench be deployed on production systems only with explicit
user consent, permission boundaries, and audit logging.

\bibliography{main}

\clearpage
\appendix

\section{Environment details}
\label{app:environment}

\subsection{Infrastructure implementation details}
\label{app:env_infra}

The design rationale for our Docker-QEMU stack is discussed in
Section~\ref{sec:env_infra}. Here we provide implementation details and a detailed comparison.

\begin{table*}[!htbp]
  \caption{Comparison of macOS agent benchmark environments.}
  \label{tab:env_compare}
  \centering
  \begin{tabular}{lccc}
    \toprule
                          & macOSWorld & macOSArena & MacAgentBench \\
    \midrule
    Infrastructure        & AWS EC2 Mac     &        Docker-QEMU stack       & Docker-QEMU stack \\
    Base image            & --              & \textasciitilde 40\,GB         & \textasciitilde 52\,GB \\
    Startup time          & \textasciitilde 15\,min & \textasciitilde 5\,min & \textasciitilde 30\,s \\
    Runtime disk usage    & --              & \textasciitilde 40\,GB         & \textasciitilde 1\,GB \\
    \bottomrule
  \end{tabular}
\end{table*}

\paragraph{Container-based isolation.}
Each task runs in its own container instance, fully isolated from other tasks.
After each evaluation, the container is discarded, guaranteeing a clean initial state
for the next task and eliminating cross-task contamination.

\paragraph{Copy-on-write accelerated startup.}
Rather than duplicating the full 52\,GB disk image for each container,
we leverage QEMU's copy-on-write mechanism to create a lightweight
differential layer on top of the base image.
The virtual machine reads directly from the shared base image and only records
modified blocks in the copy-on-write layer.
This reduces container startup time from about 5\,min (full copy) to under 30\,s,
and lowers per-instance disk overhead from 52\,GB to about 1\,GB,
with no impact on runtime performance.

\paragraph{Implementation details.}
Docker image base: sickcodes/Docker-OSX with QEMU~10.0.0 and macOS Tahoe~26.
QEMU launch parameters: 4~CPU cores, 4\,GB RAM, qcow2 disk format.
Networking: SSH on container port 10022 and VNC on container port 5900,
each dynamically mapped to host ports.
Task-level reset: between tasks, the Docker container is restarted,
discarding and rebuilding the COW layer to restore a clean initial state.
Container lifecycle: \texttt{docker run} $\rightarrow$ wait for SSH $\rightarrow$
environment initialization $\rightarrow$ execute task $\rightarrow$ evaluate $\rightarrow$ next task.
Parallel execution: up to 7 containers can run simultaneously on a single host
with 64\,GB RAM.

\subsection{Observation space}
\label{app:obs_space}

Agents receive observations through two channels:

\begin{itemize}
  \item \textbf{Screenshots}: Full desktop capture of the macOS environment,
    providing pixel-level visual information of the current GUI state.
  \item \textbf{Terminal output}: For CLI tasks, agents receive the stdout/stderr output
    from executed commands via SSH.
\end{itemize}

\subsection{Action space}
\label{app:action_space}

Agents interact with the environment through two channels:

\begin{itemize}
  \item \textbf{GUI actions}: Mouse and keyboard operations via pyautogui, including
    click (left/right/double), drag, scroll, type, and hotkey combinations.
  \item \textbf{CLI actions}: Shell commands executed via SSH, enabling file system operations,
    application launching, and AppleScript execution.
\end{itemize}

Agents can freely mix GUI and CLI actions within a single task, choosing the most efficient
interaction method for each step.

\section{Benchmark details}
\label{app:benchmark}

\subsection{Task annotation process}
\label{app:annotation}

Tasks were designed by two annotation teams composed of co-authors of
this work, with each team responsible for half of the tasks.
The annotation process consisted of three stages:
(1) team members familiarized themselves with each target application
by completing a set of tutorial tasks;
(2) each team independently designed tasks covering diverse functionality
and difficulty levels, including the instruction, setup script,
and evaluation script;
(3) after construction, the two teams swapped tasks: each team 
completed and verified the tasks designed by the other team, ensuring that
instructions are unambiguous and evaluators produce correct results.
As a final validation step, we ran the full benchmark using OpenClaw with
Claude Opus~4.6 and manually inspected the agent's execution logs for every task,
confirming that no false positives or false negatives were present
in the evaluation results.

\paragraph{Annotation team background, recruitment, compensation, and risk.}
The two annotation teams are composed of co-authors of this work
(graduate students in computer science).
Annotation was carried out as internal collaborative work on this
research project: team members were not recruited through any
crowdsourcing platform and received no hourly payment.
Their contribution to the benchmark is recognized through co-authorship
on this paper, which is the standard form of compensation for research
collaboration in academia and is independent of the team members'
country of residence.
Because all annotated content consists of benign desktop-application
tasks (e.g., notes, calendar, file operations) and contains no
offensive, politically sensitive, or psychologically demanding material,
the annotation process posed no known risk to the team members.

\subsection{Task sources}
\label{app:task_sources}

The seed tasks come from two sources.
63 tasks are adapted from macOSArena~\citeyearpar{macOSArena},
with evaluation scripts rewritten to account for macOS version changes
(see Section~\ref{sec:seed_tasks}).
The remaining 110 tasks are newly designed, covering the iWork suite
(Pages, Numbers, Keynote) and frequent desktop operations.
After cross-validation, 4 tasks with unreliable evaluators were excluded,
yielding 169 final seed tasks.

\subsection{Template-based expansion details}
\label{app:expansion}

Each of the 169 seed tasks was expanded into 4 variants through a two-stage process,
yielding $169 \times 4 = 676$ task instances.

\paragraph{Stage 1: Parameter substitution.}
We modify the concrete values in the task, such as file names, dates, content strings,
and target application states, to alter the specific behavior of the task.
The environment setup script and evaluation script are updated accordingly
to reflect the new parameter values.
For example, a task that originally creates a note titled ``Meeting Notes''
may be changed to ``Project Summary'' with different body content.

\paragraph{Stage 2: Instruction rewriting.}
We use an LLM to rephrase each task instruction while preserving the original
task objective. This produces semantically equivalent but lexically distinct
instruction variants that introduce phrasing differences across the four
variants of each seed task.
All rewritten instructions were manually reviewed to ensure semantic fidelity.

\subsection{Evaluation script examples}
\label{app:eval_examples}

Example of a shell-based evaluator (checking if a file exists at a given path):
\begin{lstlisting}
def finder_check_file_exists(env, file_path: str) -> bool:
    env.connect_ssh()
    cmd = f'test -f "{file_path}" && echo "Exists" || echo "Not found"'
    stdout, stderr = env.run_command(cmd)
    output = stdout.read().decode().strip()
        if hasattr(stdout, "read") else stdout.strip()
    return output == "Exists"
\end{lstlisting}

Example of an AppleScript evaluator (checking a reminder's due date via the Reminders app):
\begin{lstlisting}
def reminders_check_work_due_next_week(env, reminder_name="work"):
    env.connect_ssh()
    stdout, _ = env.run_command("date '+%Y-%m-%d'")
    current_date_str = stdout.read().decode().strip()
        if hasattr(stdout, "read") else stdout.strip()
    current_date = datetime.datetime.strptime(
        current_date_str, "%Y-%m-%d").date()
    current_iso_week = current_date.isocalendar()[1]

    apple_script = f"""
    tell application "Reminders"
        set workReminder to first reminder
            whose name is "{reminder_name}"
        set workDueDate to due date of workReminder
    end tell
    return workDueDate as string
    """
    stdout, stderr = env.run_command(
        f"osascript -e '{apple_script}'")
    due_str = stdout.read().decode().strip()
        if hasattr(stdout, "read") else stdout.strip()
    due_date = dateparser.parse(due_str).date()
    due_week = due_date.isocalendar()[1]
    return due_week == current_iso_week + 1
\end{lstlisting}

Example of a Python evaluator with fuzzy matching (checking live weather data against the agent's output with tolerances):
\begin{lstlisting}
def new_weather_check_contains_live_current_values(
    env, output_file: str, expected_command: str
) -> bool:
    output_text = _read_remote_file(env, output_file)
    if output_text is None:
        return False
    live_text = _run_command(env, expected_command).strip()

    live = _parse_live(live_text.splitlines()[0])
    user = _parse_user(output_text)
    if not live or not user:
        return False

    if abs(user["temp_c"] - live["temp_c"]) > 2.0:
        return False
    if abs(user["wind_kmh"] - live["wind_kmh"]) > 5.0:
        return False
    if abs(user["humidity"] - live["humidity"]) > 10.0:
        return False
    if abs(user["prec_mm"] - live["prec_mm"]) > 0.5:
        return False
    return True
\end{lstlisting}

\subsection{Task examples}
\label{app:task_examples}

We list one representative task for each of the 25 applications below, followed by an additional example illustrating a cross-application task. 
Each task is presented as a separate card.

\begin{taskbox}{Notes}
In Apple Notes, edit the note titled ``Project Ideas'' and append the line ``Reviewed on Tuesday.''.
\end{taskbox}

\begin{taskbox}{Reminders}
In the Reminders app, create a new reminder titled ``Dentist follow-up'' due ``2026-04-15 09:30''.
\end{taskbox}

\begin{taskbox}{Calendar}
In the Calendar app, create a recurring event called ``Lab Meeting'' every Monday at 10:00\,AM.
\end{taskbox}

\begin{taskbox}{Pages}
In the Pages app, create a new document using the ``Blank Black'' template and write: Passwd: 111111.
\end{taskbox}

\begin{taskbox}{Numbers}
In the Numbers app, create a new blank document, insert a table into the first sheet if needed, then enter the value ``test'' into cell A1.
\end{taskbox}

\begin{taskbox}{Keynote}
In the Keynote app, create a new document, export it to PowerPoint, and save it to the Documents folder.
\end{taskbox}

\begin{taskbox}{Obsidian}
In Obsidian, create a new note called \texttt{Inbox/Vendor Call Notes.md} and set its content exactly to the provided meeting notes.
\end{taskbox}

\begin{taskbox}{Finder}
Create a Smart Folder that shows all PDF files modified in the last 7 days and save it to \texttt{\$HOME/Library/Saved Searches} with the name \texttt{filter.savedSearch}.
\end{taskbox}

\begin{taskbox}{Terminal}
In the Terminal app, create a new Conda environment named \texttt{macos\_eval} and install the \texttt{spotify} package within that environment.
\end{taskbox}

\begin{taskbox}{Settings}
In System Settings, set the accent color to purple and turn off the option ``Tint window background with wallpaper color''.
\end{taskbox}

\begin{taskbox}{Clock}
In the Clock app, set an alarm for 6:30\,AM on weekdays (Monday through Friday) labeled ``Morning run'', using the ``Waves'' sound. Then, turn off this alarm.
\end{taskbox}

\begin{taskbox}{Safari}
Create a new folder named ``agent task'' at the top level of Safari's bookmarks (not inside the Bookmarks Bar). Then, add at least five webpages to this folder, and make sure one of them is \texttt{https://www.apple.com/}.
\end{taskbox}

\begin{taskbox}{GitHub}
Find the latest open issue in \texttt{openclaw/openclaw} with label ``bug'', and save it as \texttt{\#<number> | <labels> | <title>} to \texttt{\$HOME/Desktop/gh\_open\_issues.txt}.
\end{taskbox}

\begin{taskbox}{Himalaya}
Check the latest OTP email in Inbox (search across all pages) and save only one line to \texttt{\$HOME/Desktop/mail\_otp.txt} as \texttt{otp\_code: <6-digit-code>}. If no OTP email exists, write \texttt{otp\_code: none}.
\end{taskbox}

\begin{taskbox}{BlogWatcher}
In BlogWatcher, scan all tracked blogs for updates.
\end{taskbox}

\begin{taskbox}{Weather}
Can you check tomorrow's weather in ``San Francisco'' and tell me whether it will be rainy? Save only one yes/no line to \texttt{\$HOME/Desktop/weather\_tomorrow\_rain.txt}.
\end{taskbox}

\begin{taskbox}{VSCode}
In VS Code, install the official Python extension. Then, set the user-level conda path setting to \texttt{/opt/anaconda3/bin/conda}.
\end{taskbox}

\begin{taskbox}{Tmux}
In Terminal, send ``status'' to the tmux session ``tmux-status-send-c1'', then save the reply to \texttt{\$HOME/Desktop/tmux\_result.json}.
\end{taskbox}

\begin{taskbox}{ClawHub}
From ClawHub, install ``incident-triage-playbook'' at version ``1.0.0'' into \texttt{\$HOME/Desktop/clawhub\_skills}.
\end{taskbox}

\begin{taskbox}{SongSee}
In Terminal, use SongSee to generate a multi-panel visualization for \texttt{\$HOME/Desktop/Ode to Joy.mp3} with spectrogram, mel, chroma, and mfcc, then save it to \texttt{\$HOME/Desktop/songsee\_panel.png}.
\end{taskbox}

\begin{taskbox}{GifGrep}
Find the first Tenor GIF result for ``cat typing on keyboard'', then download that GIF and save it as \texttt{\$HOME/Desktop/gifgrep\_download.gif}.
\end{taskbox}

\begin{taskbox}{Whisper}
In Terminal, use Whisper's tiny model to make an SRT subtitle file for \texttt{\$HOME/Documents/whisper\_audio\_en.ogg} and save it to \texttt{\$HOME/Desktop/whisper\_subtitles.srt}.
\end{taskbox}

\begin{taskbox}{TTS}
In Terminal, create a spoken reminder that says ``Reminder: submit the travel reimbursement form today.'', and save it as a WAV file to \texttt{\$HOME/Desktop/sherpa\_tts.wav}.
\end{taskbox}

\begin{taskbox}{Video Frames}
In Terminal, extract a frame at 00:00:01 from \texttt{\$HOME/Desktop/benchmark\_source.mp4}, resize it to 640x720, and save it as \texttt{\$HOME/Desktop/vf\_thumb.png}.
\end{taskbox}

\begin{taskbox}{Peekaboo}
Capture a PNG screenshot of the TextEdit window and save it to \texttt{\$HOME/Desktop/peekaboo.png}.
\end{taskbox}

\begin{taskbox}{Multi-App}
In the Terminal, find the current macOS version and in the Safari app, look up the release year of that macOS version; then in the Reminders app, create a list named ``Apple Systems'' and add a reminder titled ``\{macOS Version Name\} Release'' with its due date set to that release year.
\end{taskbox}

\subsection{License and intended use}
\label{app:license}

\paragraph{Released artifacts.}
We will release MacAgentBench under permissive open-source licenses
(Apache License 2.0 for code and CC-BY 4.0 for task data and agent
trajectories). The released artifacts include: (i) task specifications
and deterministic evaluation scripts for all 676 tasks; (ii) gold
reference solutions for the subset of tasks whose solution is
expressible as CLI or AppleScript commands; (iii) full agent execution
trajectories (screenshots, actions, and checkpoint-level evaluation outcomes) for all 676 tasks; and (iv) the Docker-QEMU
infrastructure and environment initialization scripts.
The benchmark is intended primarily for non-commercial research on
evaluating CUAs and is not intended for training production-deployed
agents without additional safety review.

\paragraph{Use of existing artifacts.}
The 63 seed tasks adapted from macOSArena~\citeyearpar{macOSArena} are reused
under its original research-use license, with substantially rewritten
evaluation scripts to accommodate macOS Tahoe~26.
Our Docker-QEMU stack builds on sickcodes/Docker-OSX (GPL-3.0) and
QEMU 10.0.0 (GPL-2.0); we distribute only configuration and orchestration
code rather than redistributing macOS images, leaving the macOS image
acquisition to end users in accordance with Apple's Software License Agreement.
All evaluated models and frameworks are used through their respective
public APIs or open-source releases, in compliance with their terms of use.
We use these artifacts solely for academic evaluation purposes consistent
with their intended use.

\paragraph{Data sensitivity.}
The content in MacAgentBench comes from two sources. The majority of
content is synthetic: task instructions, file contents, and placeholder
values (e.g., names, dates, email addresses) are either fabricated by
annotation team members or generated by an LLM during template-based
expansion (Appendix~\ref{app:expansion}).
A smaller portion is adapted from team members' own personal data
(e.g., sample emails, notes, and document contents used to construct
realistic task environments), some of which originally involved third
parties such as email recipients or individuals mentioned in document
text. Before inclusion in the benchmark, all such items were manually
de-identified: real names were replaced with fabricated placeholders,
real email addresses, phone numbers, and physical addresses were
replaced with fabricated values, real file names and paths were
replaced with generic names, and any body text that could reveal
identity was rewritten.
The task domains are benign productivity workflows such as notes,
calendar, file operations, and system settings, and contain no
offensive content. During cross-validation (Appendix~\ref{app:annotation}),
the two annotation teams independently reviewed each other's tasks,
including all parameter values, file contents, and LLM-rewritten
instructions, ensuring that no personally identifying information from
either the authors or any third party remains in the benchmark.
All annotation team members, as co-authors of this work, explicitly
consent to the inclusion of their de-identified personal data in
MacAgentBench and to its release for non-commercial research use.
For content that originally involved third parties (e.g., email
recipients or individuals mentioned in document text), the
de-identification process described above removes all information
that could be associated with those individuals, so that no original
third-party identity is recoverable from the released artifacts.

\section{Experiment details}
\label{app:experiments}

\subsection{Model configurations and API pricing}
\label{app:models}

We evaluate a representative set of proprietary and open-source
vision-language models. Proprietary models are accessed through their
official APIs, while open-source models are either accessed through
OpenRouter or self-hosted on local GPUs depending on availability.
Table~\ref{tab:model_config} reports the API pricing used to compute
the cost numbers in the main paper. All prices are collected from
OpenRouter\footnote{\url{https://openrouter.ai/}}; for models without
publicly available pricing, we mark the corresponding entries as ``--''.

\paragraph{Proprietary general VLMs.}
Claude Opus~4.6~\citeyearpar{Claude4}, GPT-5.4~\citeyearpar{GPT5}, and
Gemini~3.1~Pro~\citeyearpar{Gemini3}.

\paragraph{Open-source general VLMs.}
Qwen3VL series~\citeyearpar{bai2025qwen3vltechnicalreport} (8B and 32B,
thinking variants) and InternVL3.5
series~\citeyearpar{InternVL35} (8B and 14B).

\paragraph{Native GUI agent models.}
UI-TARS~\citeyearpar{UITARS}, ScaleCUA~\citeyearpar{ScaleCUA2},
GUI-Owl-1.5~\citeyearpar{MobileAgentv35} (thinking variants), and
OpenCUA~\citeyearpar{opencua}. These models cannot reliably produce
the action formats required by AgentS3 and OpenClaw, so they are
evaluated only on the baseline framework.

\begin{table}[htbp]
  \caption{Models used in our experiments and their API pricing. Prices
  are collected from OpenRouter; ``--'' indicates that no public pricing
  is available.}
  \label{tab:model_config}
  \centering
  \small
  \begin{tabular}{lcc}
    \toprule
    Model & Input (\$/M tok) & Output (\$/M tok) \\
    \midrule
    \multicolumn{3}{l}{\textit{Proprietary}} \\
    Claude Opus 4.6      & 5.00  & 25.00 \\
    GPT-5.4              & 2.50  & 15.00 \\
    Gemini 3.1 Pro       & 2.00  & 12.00 \\
    \midrule
    \multicolumn{3}{l}{\textit{Open-source}} \\
    Qwen3VL-235B-A22B    & 0.26  & 2.60  \\
    Qwen3VL-32B          & 0.104 & 0.416 \\
    Qwen3VL-8B           & 0.117 & 1.365 \\
    OpenCUA-32B          & --    & --    \\
    OpenCUA-7B           & --    & --    \\
    GUI-Owl-1.5-32B      & --    & --    \\
    GUI-Owl-1.5-8B       & --    & --    \\
    InternVL3.5-14B         & --    & --    \\
    InternVL3.5-8B          & --    & --    \\
    UI-TARS-72B-DPO      & --    & --    \\
    UI-TARS-1.5-7B       & 0.10  & 0.20  \\
    ScaleCUA-32B         & --    & --    \\
    ScaleCUA-7B          & --    & --    \\
    \bottomrule
  \end{tabular}
\end{table}

\paragraph{Computational budget and infrastructure.}
For self-hosted open-source models (Qwen3VL series, InternVL series,
UI-TARS series, ScaleCUA series, GUI-Owl series, and OpenCUA series),
inference was performed on a cluster of NVIDIA H200 GPUs.
Model parameter counts are indicated directly in each model's name
(e.g., 235B for Qwen3VL-235B-A22B, 32B for ScaleCUA-32B,
8B for Qwen3VL-8B).
Proprietary models (Claude Opus~4.6, GPT-5.4, Gemini~3.1~Pro) were
accessed through their official APIs and required no local compute.
The total computational budget for evaluating all open-source models
across the three frameworks on the full 676-task benchmark was
approximately 8--16 NVIDIA H200 GPUs over 1--2 weeks of wall-clock time.
API costs for proprietary models are summarized per task in the Cost
column of Table~\ref{tab:main_results}, with corresponding rates listed
in Table~\ref{tab:model_config}.

\paragraph{Inference settings.}
All models are evaluated using the default sampling temperature
configured by their official API or open-source release;
we do not perform manual temperature tuning.
For consistency across models, we set a uniform context window of
262{,}144 tokens and an output budget of \texttt{max\_tokens}=16{,}384
per model call.
Each task is capped at 50 agent interaction steps, as stated in
Section~\ref{sec:exp_setup}.
Since MacAgentBench is an evaluation benchmark rather than a training
study, no hyperparameter search was performed;
results reflect each model's capability under default sampling settings.

\begin{table*}[!htbp]
  \caption{Skills in the OpenClaw skill library used in our experiments.}
  \label{tab:skill_library}
  \centering
  \footnotesize
  \begin{tabular}{p{0.18\linewidth} p{0.78\linewidth}}
    \toprule
    Skill & Description \\
    \midrule
    \texttt{1password}          & Set up and use the 1Password CLI (\texttt{op}) for sign-in, secret reading, injection, and command execution. \\
    \texttt{apple-notes}        & Manage Apple Notes via the \texttt{memo} CLI on macOS (create, view, edit, delete, search, move, and export notes). \\
    \texttt{apple-reminders}    & Manage Apple Reminders via the \texttt{remindctl} CLI: list, add, edit, complete, and delete reminders. \\
    \texttt{bear-notes}         & Create, search, and manage Bear notes via the \texttt{grizzly} CLI. \\
    \texttt{blogwatcher}        & Monitor blogs and RSS/Atom feeds for updates using the \texttt{blogwatcher} CLI. \\
    \texttt{blucli}             & BluOS CLI (\texttt{blu}) for device discovery, playback, grouping, and volume control. \\
    \texttt{bluebubbles}        & Send and manage iMessages via BlueBubbles. \\
    \texttt{camsnap}            & Capture frames or clips from RTSP/ONVIF cameras. \\
    \texttt{canvas}             & Canvas LMS operations (course and assignment access). \\
    \texttt{clawhub}            & Use the ClawHub CLI to search, install, update, and publish agent skills from clawhub.com. \\
    \texttt{coding-agent}       & Delegate coding tasks to Codex, Claude Code, or Pi agents via a background process. \\
    \texttt{discord}            & Discord operations through the generic message tool (channel=\texttt{discord}). \\
    \texttt{eightctl}           & Control Eight Sleep pods: status, temperature, alarms, and schedules. \\
    \texttt{food-order}         & Reorder Foodora orders and track ETA/status via \texttt{ordercli}. \\
    \texttt{gemini}             & Gemini CLI for one-shot Q\&A, summaries, and generation. \\
    \texttt{gh-issues}          & Fetch GitHub issues, spawn sub-agents to open PRs, and monitor/address PR review comments. \\
    \texttt{gifgrep}            & Search GIF providers with a CLI/TUI, download results, and extract stills or sheets. \\
    \texttt{github}             & GitHub operations via the \texttt{gh} CLI: issues, PRs, CI runs, code review, API queries. \\
    \texttt{gog}                & Google Workspace CLI for Gmail, Calendar, Drive, Contacts, Sheets, and Docs. \\
    \texttt{goplaces}           & Query the Google Places API (New) via the \texttt{goplaces} CLI: text search, details, and reviews. \\
    \texttt{healthcheck}        & Host security hardening and risk-tolerance configuration for OpenClaw deployments. \\
    \texttt{himalaya}           & CLI to manage emails via IMAP/SMTP: list, read, write, reply, forward, search, and organize. \\
    \texttt{imsg}               & iMessage/SMS CLI for listing chats, history, and sending messages via Messages.app. \\
    \texttt{mcporter}           & Use the \texttt{mcporter} CLI to list, configure, auth, and call MCP servers/tools (HTTP or stdio). \\
    \texttt{model-usage}        & Summarize per-model usage and cost via the CodexBar CLI for Codex or Claude. \\
    \texttt{nano-banana-pro}    & Generate or edit images via Gemini 3 Pro Image (Nano Banana Pro). \\
    \texttt{nano-pdf}           & Edit PDFs with natural-language instructions using the \texttt{nano-pdf} CLI. \\
    \texttt{notion}             & Notion API for creating and managing pages, databases, and blocks. \\
    \texttt{obsidian}           & Work with Obsidian vaults (plain Markdown notes) and automate via \texttt{obsidian-cli}. \\
    \texttt{openai-image-gen}   & Batch-generate images via the OpenAI Images API with a random prompt sampler and gallery. \\
    \texttt{openai-whisper-api} & Transcribe audio via the OpenAI Audio Transcriptions API (Whisper). \\
    \texttt{openai-whisper}     & Local speech-to-text with the Whisper CLI (no API key required). \\
    \texttt{openhue}            & Control Philips Hue lights and scenes via the OpenHue CLI. \\
    \texttt{oracle}             & Best practices for using the oracle CLI (prompt and file bundling, engines, sessions, attachments). \\
    \texttt{ordercli}           & Foodora-only CLI for checking past orders and active order status. \\
    \texttt{peekaboo}           & Capture and automate macOS UI with the Peekaboo CLI. \\
    \texttt{sag}                & ElevenLabs text-to-speech with a mac-style \texttt{say} UX. \\
    \texttt{session-logs}       & Search and analyze your own session logs (older/parent conversations) using \texttt{jq}. \\
    \texttt{sherpa-onnx-tts}    & Local text-to-speech via sherpa-onnx (offline, no cloud). \\
    \texttt{skill-creator}      & Create or update AgentSkills with scripts, references, and assets. \\
    \texttt{slack}              & Control Slack from OpenClaw via the slack tool, including reactions and pin/unpin in channels or DMs. \\
    \texttt{songsee}            & Generate spectrograms and feature-panel visualizations from audio with the \texttt{songsee} CLI. \\
    \texttt{sonoscli}           & Control Sonos speakers: discover, status, play, volume, and grouping. \\
    \texttt{spotify-player}     & Terminal Spotify playback and search via \texttt{spogo} (preferred) or \texttt{spotify\_player}. \\
    \texttt{summarize}          & Summarize or extract text/transcripts from URLs, podcasts, and local files. \\
    \texttt{things-mac}         & Manage Things 3 via the \texttt{things} CLI: add/update projects and todos, search the local database. \\
    \texttt{tmux}               & Remote-control tmux sessions for interactive CLIs by sending keystrokes and scraping pane output. \\
    \texttt{trello}             & Manage Trello boards, lists, and cards via the Trello REST API. \\
    \texttt{video-frames}       & Extract frames or short clips from videos using \texttt{ffmpeg}. \\
    \texttt{voice-call}         & Start voice calls via the OpenClaw voice-call plugin. \\
    \texttt{wacli}              & Send WhatsApp messages or search/sync WhatsApp history via the \texttt{wacli} CLI. \\
    \texttt{weather}            & Get current weather and forecasts via wttr.in or Open-Meteo. \\
    \texttt{xurl}               & Authenticated requests to the X (Twitter) API: post, reply, quote, search, DMs, and media upload. \\
    \bottomrule
  \end{tabular}
\end{table*}

\subsection{Agent framework configurations}
\label{app:frameworks}

All three frameworks share the same environment interface and a budget
of 50 interaction steps per task. They differ in their action spaces
and the external tools exposed to the agent. The complete system
prompts and user message templates for each framework are provided in
Appendix~\ref{app:prompts}.

\paragraph{Baseline (/).}
A pure GUI agent that interacts with the environment through
screenshots and \texttt{pyautogui} code output. Claude Opus~4.6,
GPT-5.4, and Gemini~3.1~Pro share the same \texttt{pyautogui}-based
action space and prompt template.

\paragraph{AgentS3.}
We reuse the official multi-agent architecture of
AgentS3~\citeyearpar{AgentS3}, which consists of a planner, reflector, coder,
and grounder, and extends the baseline GUI action space with
AppleScript and code execution.

\paragraph{OpenClaw.}
OpenClaw~\citeyearpar{openclaw2025} is deployed inside the container as a
native macOS application and, on top of the action space of AgentS3,
further provides tools such as \texttt{Skill}
(Appendix~\ref{app:skills}), \texttt{Memory}, and \texttt{WebSearch}.

\subsection{Skill library}
\label{app:skills}

Our skills are taken from the full set of skills that ship with the
official OpenClaw download. We install the required dependencies and
activate all of them, so that the agent has access to every shipped
skill during evaluation. Table~\ref{tab:skill_library} lists these
skills together with a brief description.

\subsection{Prompt templates}
\label{app:prompts}

We present the prompt templates used for each of the three frameworks,
including system prompts and user message formats. Prompts 1a/1b/1c/1d/1e/1f
and Prompt 2 are reproduced from the official agent
implementations; Prompt 3 is shown with two simplifications described
below.

\subsubsection{Baseline (/)}

\hspace{\parindent}Prompt 1a is shared by Claude Opus~4.6, GPT-5.4, and Gemini~3.1~Pro.
All three models are prompted with the same pyautogui-based action space and the
same input/output format.

\begin{tcolorbox}[enhanced jigsaw, enforce breakable, pad at break=2mm, colback=blue!3!white, colframe=blue!50!black, left=1mm, right=1mm, top=1mm, bottom=1mm, title=Prompt 1a: Baseline -- Claude Opus 4.6 / GPT-5.4 / Gemini 3.1 Pro]
\small
\tcbsubtitle{System Prompt}
\begin{lstlisting}[language=,basicstyle=\ttfamily\scriptsize,breaklines=true,breakindent=0pt,frame=none,backgroundcolor=\color{blue!3!white},xleftmargin=0pt,xrightmargin=0pt,aboveskip=2pt,belowskip=2pt]
You are an agent which follow my instruction and perform desktop computer tasks as instructed.
You have good knowledge of computer and good internet connection and assume your code will run on a computer for controlling the mouse and keyboard.
For each step, you will get an observation of an image, which is the screenshot of the computer screen and you will predict the action of the computer based on the image.

You are required to use `pyautogui` to perform the action grounded to the observation, but DONOT use the `pyautogui.locateCenterOnScreen` function to locate the element you want to operate with since we have no image of the element you want to operate with. DONOT USE `pyautogui.screenshot()` to make screenshot. Under macOS, the `pyautogui.rightClick` function may not work properly due to a known bug in `pyautogui` itself. In such cases, consider using the `pynput` library as an alternative specifically for implementing right-click functionality.
Return one line or multiple lines of python code to perform the action each time, be time efficient. When predicting multiple lines of code, make some small sleep like `time.sleep(0.5);` interval so that the machine could take; Each time you need to predict a complete code, no variables or function can be shared from history
You need to specify the coordinates of by yourself based on your observation of current observation, but you should be careful to ensure that the coordinates are correct.
You ONLY need to return the code inside a code block, like this:
```python
# your code here
```
Specially, it is also allowed to return the following special code:
When you think you have to wait for some time, return ```WAIT```;
When you think the task can not be done, return ```FAIL```, don't easily say ```FAIL```, try your best to do the task;
When you think the task is done, return ```DONE```.

My computer's password is {password}, feel free to use it when you need sudo rights.
First give the current screenshot and previous things we did a short reflection, then RETURN ME THE CODE OR SPECIAL CODE I ASKED FOR. NEVER EVER RETURN ME ANYTHING ELSE.

You are asked to complete the following task: {instruction}
\end{lstlisting}

\tcbsubtitle{User Message}
\begin{lstlisting}[language=,basicstyle=\ttfamily\scriptsize,breaklines=true,breakindent=0pt,frame=none,backgroundcolor=\color{blue!3!white},xleftmargin=0pt,xrightmargin=0pt,aboveskip=2pt,belowskip=2pt]
Given the screenshot as below. What's the next step that you will do to help with the task?
<screenshot>
\end{lstlisting}
\end{tcolorbox}

Prompt 1b is used for Qwen3VL models, which adopt a tool-calling format with
a built-in \texttt{computer\_use} tool schema and \texttt{<tool\_call>} response tags.

\begin{tcolorbox}[enhanced jigsaw, enforce breakable, pad at break=2mm, colback=blue!3!white, colframe=blue!50!black, left=1mm, right=1mm, top=1mm, bottom=1mm, title=Prompt 1b: Baseline -- Qwen3VL]
\small
\tcbsubtitle{System Prompt}
\begin{lstlisting}[language=,basicstyle=\ttfamily\scriptsize,breaklines=true,breakindent=0pt,frame=none,backgroundcolor=\color{blue!3!white},xleftmargin=0pt,xrightmargin=0pt,aboveskip=2pt,belowskip=2pt]
# Tools

You may call one or more functions to assist with the user query.

You are provided with function signatures within <tools></tools> XML tags:
<tools>
{
  "type": "function",
  "function": {
    "name_for_human": "computer_use",
    "name": "computer_use",
    "description": "Use a mouse and keyboard to interact with a computer, and take screenshots.\n* This is an interface to a desktop GUI. You do not have access to a terminal or applications menu. You must click on desktop icons to start applications.\n* Some applications may take time to start or process actions, so you may need to wait and take successive screenshots to see the results of your actions. E.g. if you click on Firefox and a window doesn't open, try wait and taking another screenshot.\n* The screen's resolution is {processed_width}x{processed_height}.\n* Whenever you intend to move the cursor to click on an element like an icon, you should consult a screenshot to determine the coordinates of the element before moving the cursor.\n* If you tried clicking on a program or link but it failed to load even after waiting, try adjusting your cursor position so that the tip of the cursor visually falls on the element that you want to click.\n* Make sure to click any buttons, links, icons, etc with the cursor tip in the center of the element. Don't click boxes on their edges unless asked.",
    "parameters": {
      "properties": {
        "action": {
          "description": "* `key`: Performs key down presses on the arguments passed in order, then performs key releases in reverse order.\n* `type`: Type a string of text on the keyboard.\n* `mouse_move`: Move the cursor to a specified (x, y) pixel coordinate on the screen.\n* `left_click`: Click the left mouse button at a specified (x, y) pixel coordinate on the screen.\n* `left_click_drag`: Click and drag the cursor to a specified (x, y) pixel coordinate on the screen.\n* `right_click`: Click the right mouse button at a specified (x, y) pixel coordinate on the screen.\n* `middle_click`: Click the middle mouse button at a specified (x, y) pixel coordinate on the screen.\n* `double_click`: Double-click the left mouse button at a specified (x, y) pixel coordinate on the screen.\n* `triple_click`: Triple-click the left mouse button at a specified (x, y) pixel coordinate on the screen (simulated as double-click since it's the closest action).\n* `scroll`: Performs a scroll of the mouse scroll wheel.\n* `hscroll`: Performs a horizontal scroll (mapped to regular scroll).\n* `wait`: Wait specified seconds for the change to happen.\n* `terminate`: Terminate the current task and report its completion status.\n* `answer`: Answer a question.",
          "enum": ["key", "type", "mouse_move", "left_click", "left_click_drag", "right_click", "middle_click", "double_click", "scroll", "wait", "terminate"],
          "type": "string"
        },
        "keys":       {"description": "Required only by `action=key`.",  "type": "array"},
        "text":       {"description": "Required only by `action=type`.", "type": "string"},
        "coordinate": {"description": "The x,y coordinates for mouse actions.", "type": "array"},
        "pixels":     {"description": "The amount of scrolling.", "type": "number"},
        "time":       {"description": "The seconds to wait.", "type": "number"},
        "status":     {"description": "The status of the task.", "type": "string", "enum": ["success", "failure"]}
      },
      "required": ["action"],
      "type": "object"
    },
    "args_format": "Format the arguments as a JSON object."
  }
}
</tools>

For each function call, return a json object with function name and arguments within <tool_call></tool_call> XML tags:
<tool_call>
{"name": <function-name>, "arguments": <args-json-object>}
</tool_call>

# Response format

Response format for every step:
1) Action: a short imperative describing what to do in the UI.
2) A single <tool_call>...</tool_call> block containing only the JSON: {"name": <function-name>, "arguments": <args-json-object>}.

Rules:
- Output exactly in the order: Action, <tool_call>.
- Be brief: one sentence for Action.
- Do not output anything else outside those parts.
- If finishing, use action=terminate in the tool call.
\end{lstlisting}

\tcbsubtitle{User Message}
\begin{lstlisting}[language=,basicstyle=\ttfamily\scriptsize,breaklines=true,breakindent=0pt,frame=none,backgroundcolor=\color{blue!3!white},xleftmargin=0pt,xrightmargin=0pt,aboveskip=2pt,belowskip=2pt]
Please generate the next move according to the UI screenshot, instruction and previous actions.

Instruction: {instruction}

Previous actions:
{previous_actions}
\end{lstlisting}
\end{tcolorbox}

Prompt 1c is shared by InternVL and ScaleCUA, which follow the same
pyautogui-based action space with structured think/operation/action output.

\begin{tcolorbox}[enhanced jigsaw, enforce breakable, pad at break=2mm, colback=blue!3!white, colframe=blue!50!black, left=1mm, right=1mm, top=1mm, bottom=1mm, title=Prompt 1c: Baseline -- InternVL / ScaleCUA]
\small
\tcbsubtitle{System Prompt}
\begin{lstlisting}[language=,basicstyle=\ttfamily\scriptsize,breaklines=true,breakindent=0pt,frame=none,backgroundcolor=\color{blue!3!white},xleftmargin=0pt,xrightmargin=0pt,aboveskip=2pt,belowskip=2pt]
You are an autonomous GUI agent operating on the **macOS** platform. Your primary function is to analyze screen captures and perform appropriate UI actions to complete assigned tasks.

## Action Space

def click(
    x: float | None = None,
    y: float | None = None,
    clicks: int = 1,
    button: str = "left",
) -> None:
    """Clicks on the screen at the specified coordinates. The `x` and `y` parameter specify where the mouse event occurs. If not provided, the current mouse position is used. The `clicks` parameter specifies how many times to click, and the `button` parameter specifies which mouse button to use ('left', 'right', or 'middle')."""
    pass


def doubleClick(
    x: float | None = None,
    y: float | None = None,
    button: str = "left",
) -> None:
    """Performs a double click. This is a wrapper function for click(x, y, 2, 'left')."""
    pass


def rightClick(x: float | None = None, y: float | None = None) -> None:
    """Performs a right mouse button click. This is a wrapper function for click(x, y, 1, 'right')."""
    pass


def scroll(clicks: int, x: float | None = None, y: float | None = None) -> None:
    """Performs a scroll of the mouse scroll wheel at the specified coordinates. The `clicks` specifies how many clicks to scroll. The direction of the scroll (vertical or horizontal) depends on the underlying operating system. Normally, positive values scroll up, and negative values scroll down."""
    pass


def moveTo(x: float, y: float) -> None:
    """Move the mouse to the specified coordinates."""
    pass


def dragTo(
    x: float | None = None, y: float | None = None, button: str = "left"
) -> None:
    """Performs a drag-to action with optional `x` and `y` coordinates and button."""
    pass


def press(keys: str | list[str], presses: int = 1) -> None:
    """Performs a keyboard key press down, followed by a release. The function supports pressing a single key or a list of keys, multiple presses, and customizable intervals between presses."""
    pass


def hotkey(*args: str) -> None:
    """Performs key down presses on the arguments passed in order, then performs key releases in reverse order. This is used to simulate keyboard shortcuts (e.g., 'Ctrl-Shift-C')."""
    pass


def keyDown(key: str) -> None:
    """Performs a keyboard key press without the release. This will put that key in a held down state."""
    pass


def keyUp(key: str) -> None:
    """Performs a keyboard key release (without the press down beforehand)."""
    pass


def write(message: str) -> None:
    """Write the specified text."""
    pass


def call_user() -> None:
    """Call the user."""
    pass


def wait(seconds: int = 3) -> None:
    """Wait for the change to happen."""
    pass


def response(answer: str) -> None:
    """Answer a question or provide a response to an user query."""
    pass


def terminate(status: str = "success", info: str | None = None) -> None:
    """Terminate the current task with a status. The `status` specifies the termination status ('success', 'failure'), and the `info` can provide additional information about the termination."""
    pass

## Input Specification
- Screenshot of the current screen + task description + your past interaction history with UI to finish assigned tasks.

## Output Format
```
<think>
[Your reasoning process here]
</think>
<operation>
[Next intended operation description]
</operation>
<action>
[A set of executable action command]
</action>
```

## Note
- Avoid actions that would lead to invalid states.
- The generated action(s) must exist within the defined action space.
- The reasoning process, operation and action(s) in your response should be enclosed within <think></think>, <operation></operation> and <action></action> tags, respectively.
\end{lstlisting}

\tcbsubtitle{User Message}
\begin{lstlisting}[language=,basicstyle=\ttfamily\scriptsize,breaklines=true,breakindent=0pt,frame=none,backgroundcolor=\color{blue!3!white},xleftmargin=0pt,xrightmargin=0pt,aboveskip=2pt,belowskip=2pt]
<image>
Please generate the next move according to the UI screenshot, task and previous operations.

Task:
{instruction}
\end{lstlisting}
\end{tcolorbox}

Prompt 1d is used for UI-TARS models, which follow a Thought/Action output
format native to the UI-TARS architecture and bake the action space directly
into the user message.

\begin{tcolorbox}[enhanced jigsaw, enforce breakable, pad at break=2mm, colback=blue!3!white, colframe=blue!50!black, left=1mm, right=1mm, top=1mm, bottom=1mm, title=Prompt 1d: Baseline -- UI-TARS]
\small
\tcbsubtitle{System Prompt}
\begin{lstlisting}[language=,basicstyle=\ttfamily\scriptsize,breaklines=true,breakindent=0pt,frame=none,backgroundcolor=\color{blue!3!white},xleftmargin=0pt,xrightmargin=0pt,aboveskip=2pt,belowskip=2pt]
You are a helpful assistant.
\end{lstlisting}

\tcbsubtitle{User Message}
\begin{lstlisting}[language=,basicstyle=\ttfamily\scriptsize,breaklines=true,breakindent=0pt,frame=none,backgroundcolor=\color{blue!3!white},xleftmargin=0pt,xrightmargin=0pt,aboveskip=2pt,belowskip=2pt]
You are a GUI agent. You are given a task and your action history, with screenshots. You need to perform the next action to complete the task.

## Output Format
```
Thought: ...
Action: ...
```

## Action Space
click(start_box='<|box_start|>(x1,y1)<|box_end|>')
left_double(start_box='<|box_start|>(x1,y1)<|box_end|>')
right_single(start_box='<|box_start|>(x1,y1)<|box_end|>')
drag(start_box='<|box_start|>(x1,y1)<|box_end|>', end_box='<|box_start|>(x3,y3)<|box_end|>')
hotkey(key='')
type(content='') #If you want to submit your input, use "\n" at the end of `content`.
scroll(start_box='<|box_start|>(x1,y1)<|box_end|>', direction='down or up or right or left')
wait() #Sleep for 5s and take a screenshot to check for any changes.
finished(content='xxx') # Use escape characters \', \", and \n in content part to ensure we can parse the content in normal python string format.

## Note
- Use English in `Thought` part.
- Write a small plan and finally summarize your next action (with its target element) in one sentence in `Thought` part.

## User Instruction
{instruction}
\end{lstlisting}
\end{tcolorbox}

Prompt 1e is used for OpenCUA-32B and OpenCUA-7B, which follow a structured
output format with explicit terminate semantics.

\begin{tcolorbox}[enhanced jigsaw, enforce breakable, pad at break=2mm, colback=blue!3!white, colframe=blue!50!black, left=1mm, right=1mm, top=1mm, bottom=1mm, title=Prompt 1e: Baseline -- OpenCUA]
\small
\tcbsubtitle{System Prompt}
\begin{lstlisting}[language=,basicstyle=\ttfamily\scriptsize,breaklines=true,breakindent=0pt,frame=none,backgroundcolor=\color{blue!3!white},xleftmargin=0pt,xrightmargin=0pt,aboveskip=2pt,belowskip=2pt]
You are a GUI agent. You are given a task and a screenshot of the screen. You need to perform a series of pyautogui actions to complete the task. The password of the computer is "{password}". If the task is not possible to do, output the action computer.terminate(status='failure').

For each step, provide your response in this format:

Thought:
  - Step by Step Progress Assessment:
    - Analyze completed task parts and their contribution to the overall goal
    - Reflect on potential errors, unexpected results, or obstacles
    - If previous action was incorrect, predict a logical recovery step
  - Next Action Analysis:
    - List possible next actions based on current state
    - Evaluate options considering current state and previous actions
    - Propose most logical next action
    - Anticipate consequences of the proposed action
  - For Text Input Actions:
    - Note current cursor position
    - Consolidate repetitive actions (specify count for multiple keypresses)
    - Describe expected final text outcome
  - Use first-person perspective in reasoning

Action:
  Provide clear, concise, and actionable instructions:
  - If the action involves interacting with a specific target:
    - Describe target explicitly without using coordinates
    - Specify element names when possible (use original language if non-English)
    - Describe features (shape, color, position) if name unavailable
    - For window control buttons, identify correctly (minimize "-", maximize "[]", close "X")
  - if the action involves keyboard actions like 'press', 'write', 'hotkey':
    - Consolidate repetitive keypresses with count
    - Specify expected text outcome for typing actions

Finally, output the action as PyAutoGUI code or the following functions:
- {"name": "computer.triple_click", "description": "Triple click on the screen", "parameters": {"type": "object", "properties": {"x": {"type": "number", "description": "The x coordinate of the triple click"}, "y": {"type": "number", "description": "The y coordinate of the triple click"}}, "required": ["x", "y"]}}
- {"name": "computer.terminate", "description": "Terminate the current task and report its completion status", "parameters": {"type": "object", "properties": {"status": {"type": "string", "enum": ["success", "failure"], "description": "The status of the task"}}, "required": ["status"]}}
\end{lstlisting}

\tcbsubtitle{User Message}
\begin{lstlisting}[language=,basicstyle=\ttfamily\scriptsize,breaklines=true,breakindent=0pt,frame=none,backgroundcolor=\color{blue!3!white},xleftmargin=0pt,xrightmargin=0pt,aboveskip=2pt,belowskip=2pt]
# Task Instruction:
{instruction}

Please generate the next move according to the screenshot, task instruction and previous steps (if provided).
\end{lstlisting}
\end{tcolorbox}


Prompt 1f is used for GUI-Owl-1.5-32B and GUI-Owl-1.5-8B. 

\begin{tcolorbox}[enhanced jigsaw, enforce breakable, pad at break=2mm, colback=blue!3!white, colframe=blue!50!black, left=1mm, right=1mm, top=1mm, bottom=1mm, title=Prompt 1f: Baseline -- GUI-Owl-1.5]
\small
\tcbsubtitle{System Prompt}
\begin{lstlisting}[language=,basicstyle=\ttfamily\scriptsize,breaklines=true,breakindent=0pt,frame=none,backgroundcolor=\color{blue!3!white},xleftmargin=0pt,xrightmargin=0pt,aboveskip=2pt,belowskip=2pt]
# Tools

You may call one or more functions to assist with the user query.

You are provided with function signatures within <tools></tools> XML tags:
<tools>
{"type": "function", "function": {"name": "computer_use", "description": "Use a mouse and keyboard to interact with a computer, and take screenshots.\n* This is an interface to a desktop GUI. You do not have access to a terminal or applications menu. You must click on desktop icons to start applications.\n* Some applications may take time to start or process actions, so you may need to wait and take successive screenshots to see the results of your actions. E.g. if you click on Firefox and a window doesn't open, try wait and taking another screenshot.\n* The screen's resolution is 1000x1000.\n* Make sure to click any buttons, links, icons, etc with the cursor tip in the center of the element. Don't click boxes on their edges unless asked.", "parameters": {"properties": {"action": {"description": "The action to perform. The available actions are:\n* `key`: Performs key down presses on the arguments passed in order, then performs key releases in reverse order.\n* `type`: Type a string of text on the keyboard.\n* `mouse_move`: Move the cursor to a specified (x, y) pixel coordinate on the screen.\n* `left_click`: Click the left mouse button at a specified (x, y) pixel coordinate on the screen.\n* `left_click_drag`: Click and drag the cursor to a specified (x, y) pixel coordinate on the screen.\n* `right_click`: Click the right mouse button at a specified (x, y) pixel coordinate on the screen.\n* `middle_click`: Click the middle mouse button at a specified (x, y) pixel coordinate on the screen.\n* `double_click`: Double-click the left mouse button at a specified (x, y) pixel coordinate on the screen.\n* `triple_click`: Triple-click the left mouse button at a specified (x, y) pixel coordinate on the screen.\n* `scroll`: Performs a scroll of the mouse scroll wheel.\n* `hscroll`: Performs a horizontal scroll.\n* `wait`: Wait specified seconds for the change to happen.\n* `terminate`: Terminate the current task and report its completion status.\n* `answer`: Answer a question.\n* `interact`: Resolve the blocking window by interacting with the user.", "enum": ["key", "type", "mouse_move", "left_click", "left_click_drag", "right_click", "middle_click", "double_click", "triple_click", "scroll", "hscroll", "wait", "terminate", "answer", "interact"], "type": "string"}, "keys": {"description": "Required only by `action=key`.", "type": "array"}, "text": {"description": "Required only by `action=type`, `action=answer` and `action=interact`.", "type": "string"}, "coordinate": {"description": "(x, y): The x (pixels from the left edge) and y (pixels from the top edge) coordinates to move the mouse to. Required only by `action=mouse_move` and `action=left_click_drag`.", "type": "array"}, "pixels": {"description": "The amount of scrolling to perform. Positive values scroll up, negative values scroll down. Required only by `action=scroll` and `action=hscroll`.", "type": "number"}, "time": {"description": "The seconds to wait. Required only by `action=wait`.", "type": "number"}, "status": {"description": "The status of the task. Required only by `action=terminate`.", "type": "string", "enum": ["success", "failure"]}}, "required": ["action"], "type": "object"}}}
</tools>

For each function call, return a json object with function name and arguments within <tool_call></tool_call> XML tags:
<tool_call>
{"name": <function-name>, "arguments": <args-json-object>}
</tool_call>

# Response format

Response format for every step:
1) Action: a short imperative describing what to do in the UI.
2) A single <tool_call>...</tool_call> block containing only the JSON: {"name": <function-name>, "arguments": <args-json-object>}.

Rules:
- Output exactly in the order: Action, <tool_call>.
- Be brief: one for Action.
- Do not output anything else outside those two parts.
- If finishing, use action=terminate in the tool call.
\end{lstlisting}

\tcbsubtitle{User Message}
\begin{lstlisting}[language=,basicstyle=\ttfamily\scriptsize,breaklines=true,breakindent=0pt,frame=none,backgroundcolor=\color{blue!3!white},xleftmargin=0pt,xrightmargin=0pt,aboveskip=2pt,belowskip=2pt]
Please generate the next move according to the UI screenshot, instruction and previous actions.

Instruction: {instruction}

Previous actions:
{previous_actions}
\end{lstlisting}
\end{tcolorbox}

\subsubsection{AgentS3}

AgentS3 mainly consists of the grounding module, worker module, code agent module, and reflection module. In our experiments, the grounding model uses a fixed UI-TARS-1.5-7B model. Additionally, no BehaviorJudge module was introduced to implement test-time scaling.

\textbf{Grounding}

The grounding module is responsible for converting high-level descriptions into screen coordinates. One approach uses a visual grounding model, while the other uses an OCR vocabulary combined with a text-span model. Prompt 2a is employed for the visual grounding model, which converts control descriptions into coordinates that are subsequently used for actions such as click or type. Prompt 2b is used for the OCR vocabulary + text-span model, which is primarily applied to select a specific segment of text.

\begin{tcolorbox}[enhanced jigsaw, enforce breakable, pad at break=2mm, colback=green!3!white, colframe=green!50!black, left=1mm, right=1mm, top=1mm, bottom=1mm, title=Prompt 2a: GUI Grounding Model Prompt]
\small
\tcbsubtitle{User Message}
\begin{lstlisting}[language=,basicstyle=\ttfamily\scriptsize,breaklines=true,breakindent=0pt,frame=none,backgroundcolor=\color{green!3!white},xleftmargin=0pt,xrightmargin=0pt,aboveskip=2pt,belowskip=2pt]
Query:{ref_expr}
Output only the coordinate of one point in your response.
Image width: {grounding_width}, height: {grounding_height}
\end{lstlisting}
\end{tcolorbox}

\begin{tcolorbox}[enhanced jigsaw, enforce breakable, pad at break=2mm, colback=green!3!white, colframe=green!50!black, left=1mm, right=1mm, top=1mm, bottom=1mm, title=Prompt 2b: OCR Model Prompt]
\small
\tcbsubtitle{System Prompt}
\begin{lstlisting}[language=,basicstyle=\ttfamily\scriptsize,breaklines=true,breakindent=0pt,frame=none,backgroundcolor=\color{green!3!white},xleftmargin=0pt,xrightmargin=0pt,aboveskip=2pt,belowskip=2pt]
You are an expert in graphical user interfaces. Your task is to process a phrase of text, and identify the most relevant word on the computer screen.
You are provided with a phrase, a table with all the text on the screen, and a screenshot of the computer screen. You will identify the single word id that is best associated with the provided phrase.
This single word must be displayed on the computer screenshot, and its location on the screen should align with the provided phrase.
Each row in the text table provides 2 pieces of data in the following order. 1st is the unique word id. 2nd is the corresponding word.

To be successful, it is very important to follow all these rules:
1. First, think step by step and generate your reasoning about which word id to click on.
2. Then, output the unique word id. Remember, the word id is the 1st number in each row of the text table.
3. If there are multiple occurrences of the same word, use the surrounding context in the phrase to choose the correct one. Pay very close attention to punctuation and capitalization.
\end{lstlisting}
\tcbsubtitle{User Message}
\begin{lstlisting}[language=,basicstyle=\ttfamily\scriptsize,breaklines=true,breakindent=0pt,frame=none,backgroundcolor=\color{green!3!white},xleftmargin=0pt,xrightmargin=0pt,aboveskip=2pt,belowskip=2pt]
**Important**: Output the word id of the FIRST word in the provided phrase.
Phrase: {phrase}
{ocr_table}
\end{lstlisting}
\tcbsubtitle{User Message}
\begin{lstlisting}[language=,basicstyle=\ttfamily\scriptsize,breaklines=true,breakindent=0pt,frame=none,backgroundcolor=\color{green!3!white},xleftmargin=0pt,xrightmargin=0pt,aboveskip=2pt,belowskip=2pt]
Screenshot:
{screenshot}
\end{lstlisting}
\end{tcolorbox}

\textbf{Planning}

The planning model primarily observes the task context along with the latest screenshot and determines the next action. Prompt 2c is used to provide the planning model with capability boundaries, specifying which sub-agents it is allowed to utilize.

\begin{tcolorbox}[enhanced jigsaw, enforce breakable, pad at break=2mm, colback=green!3!white, colframe=green!50!black, left=1mm, right=1mm, top=1mm, bottom=1mm, title=Prompt 2c: GUI Policy Prompt]
\small
\tcbsubtitle{System Prompt}
\begin{lstlisting}[language=,basicstyle=\ttfamily\scriptsize,breaklines=true,breakindent=0pt,frame=none,backgroundcolor=\color{green!3!white},xleftmargin=0pt,xrightmargin=0pt,aboveskip=2pt,belowskip=2pt]
You are an expert in graphical user interfaces and Python code. You are responsible for executing the task: `TASK_DESCRIPTION`.
You are working in CURRENT_OS.

# GUIDELINES

## Agent Usage Guidelines
You have access to both GUI and code agents. Choose the appropriate agent based on the task requirements:

### GUI Agent
- **Use for**: clicking, typing, navigation, file operations, tasks requiring specific application features, visual elements, interactive features, application UI, complex formatting, print/export settings, multi-step workflows, pivot tables, charts

### Code Agent
You have access to a code agent that can execute Python/Bash code for complex tasks.

Use code agent for:
- **ALL spreadsheet calculations**: sums, totals, averages, formulas, data filling, missing value calculations
- **ALL data manipulation tasks**: including calculations, data processing (filtering, sorting, replacing, cleanup), bulk operations (filling or transforming ranges), formatting changes (number/date/currency formats, styles), and large-scale data entry or editing

**Usage Strategy**:
- **Full Task**: Use `agent.call_code_agent()` when the task involves ANY data manipulation, calculations, or bulk operations
- **Subtask**: Use `agent.call_code_agent("specific subtask")` for focused data tasks
- **CRITICAL**: If calling the code agent for the full task, pass the original task instruction without rewording or modification

### Code Agent Result Interpretation
- The code agent runs Python/Bash code in the background (up to 20 steps), independently performing tasks like file modification, package installation, or system operations.
- After execution, you receive a report with:
    * Steps completed (actual steps run)
    * Max steps (step budget)
    * Completion reason: DONE (success), FAIL (gave up), or BUDGET_EXHAUSTED (used all steps)
    * Summary of work done
    * Full execution history
- Interpretation:
    * DONE: The code agent finished before using all steps, believing the task was completed through code.
    * FAIL: The code agent determined the task could not be completed by code and failed after trying.
    * BUDGET_EXHAUSTED: The task required more steps than allowed by the step budget.

### Code Agent Verification
- After the code agent modifies files, your job is to find and verify these files via GUI actions (e.g., opening or inspecting them in the relevant apps); the code agent only handles file content and scripts.
- ALWAYS verify code agent results with GUI actions before using agent.done(); NEVER trust code agent output alone. If verification or the code agent fails, use GUI actions to finish the task and only use agent.done() if results match expectations.
- **CRITICAL**: Files modified by code agent may not show changes in currently open applications - you MUST close and reopen the entire application. Reloading the page/file is insufficient.

# General Task Guidelines
- For formatting tasks, always use the code agent for proper formatting.
- **Never use the code agent for charts, graphs, pivot tables, or visual elements--always use the GUI for those.**
- If creating a new sheet with no name specified, use default sheet names (e.g., "Sheet1", "Sheet2", etc.).
- After opening or reopening applications, wait at least 3 seconds for full loading.
- Don't provide specific row/column numbers to the coding agent; let it infer the spreadsheet structure itself.

Never assume a task is done based on appearances-always ensure the specific requested action has been performed and verify the modification. If you haven't executed any actions, the task is not complete.

### END OF GUIDELINES

You are provided with:
1. A screenshot of the current time step.
2. The history of your previous interactions with the UI.
3. Access to the following class and methods to interact with the UI:
class Agent:
{dynamic_api}
Your response should be formatted like this:
(Previous action verification)
Carefully analyze based on the screenshot if the previous action was successful. If the previous action was not successful, provide a reason for the failure.

(Screenshot Analysis)
Closely examine and describe the current state of the desktop along with the currently open applications.

(Next Action)
Based on the current screenshot and the history of your previous interaction with the UI, decide on the next action in natural language to accomplish the given task.

(Grounded Action)
Translate the next action into code using the provided API methods. Format the code like this:
```python
agent.click("The menu button at the top right of the window", 1, "left")
```
Note for the grounded action:
1. Only perform one action at a time.
2. Do not put anything other than python code in the block. You can only use one function call at a time. Do not put more than one function call in the block.
3. You must use only the available methods provided above to interact with the UI, do not invent new methods.
4. Only return one code block every time. There must be a single line of code in the code block.
5. Do not do anything other than the exact specified task. Return with `agent.done()` immediately after the subtask is completed or `agent.fail()` if it cannot be completed.
6. Whenever possible, your grounded action should use hot-keys with the agent.hotkey() action instead of clicking or dragging.
7. My computer's password is '{password}', feel free to use it when you need sudo rights.
8. Generate agent.fail() as your grounded action if you get exhaustively stuck on the task and believe it is impossible.
9. Generate agent.done() as your grounded action when your believe the task is fully complete.
10. Do not use the "command" + "tab" hotkey on MacOS.
11. Prefer hotkeys and application features over clicking on text elements when possible. Highlighting text is fine.
\end{lstlisting}
\tcbsubtitle{User Message}
\begin{lstlisting}[language=,basicstyle=\ttfamily\scriptsize,breaklines=true,breakindent=0pt,frame=none,backgroundcolor=\color{green!3!white},xleftmargin=0pt,xrightmargin=0pt,aboveskip=2pt,belowskip=2pt]
{initial_note_if_turn_0}
{optional_reflection_block}
Current Text Buffer = [{note1,note2,...}]
{optional_code_agent_result_block}
\end{lstlisting}
\end{tcolorbox}

\textbf{Code Agent}

The Code Agent is a separately abstracted module within AgentS3, responsible for generating and executing code. Prompt 2d is used to inform the Code Agent of its constraints and the required code format.

\begin{tcolorbox}[enhanced jigsaw, enforce breakable, pad at break=2mm, colback=green!3!white, colframe=green!50!black, left=1mm, right=1mm, top=1mm, bottom=1mm, title=Prompt 2d: Code Agent Prompt]
\small
\tcbsubtitle{System Prompt}
\begin{lstlisting}[language=,basicstyle=\ttfamily\scriptsize,breaklines=true,breakindent=0pt,frame=none,backgroundcolor=\color{green!3!white},xleftmargin=0pt,xrightmargin=0pt,aboveskip=2pt,belowskip=2pt]
You are a code execution agent with a limited step budget to complete tasks.

# Core Guidelines:
- Execute Python/Bash code step-by-step to progress toward the goal
- Use sudo with: "echo {password} | sudo -S [COMMANDS]"
- Username: "user"
- Print results and handle errors appropriately
- Code execution may not show immediately on screen

# CRITICAL: Incremental Step-by-Step Approach
- Break down complex tasks into small, self-contained steps
- Each step should contain a single, focused code snippet that advances toward the goal
- Code from each step does NOT persist to the next step - write complete, standalone snippets
- Example workflow:
    * Step 1: Write code to locate/find the target file
    * Step 2: Write code to **THOROUGHLY** inspect/read the file contents
    * Step 3: Write code to modify the file based on findings
    * Step 4: Write code to verify the changes
    - If verification fails (the modification did not work as intended), return to Step 3 and rewrite the modification code. Repeat until verification succeeds.
- Do NOT write entire scripts in one step - focus on one small task per step

# CRITICAL: Data Format Guidelines
- Store dates as proper date objects, not text strings
- Store numbers as numeric values, not formatted text with symbols
- Preserve data types for calculations and evaluations
- When applying data validation to spreadsheet columns, limit the range to only the rows containing actual data, not entire columns
- When creating cross-sheet references, use cell references (e.g., =Sheet1!A1) instead of manually typing values
- When asked to create a new sheet and no specific name is provided, default to the default sheet name (e.g., "Sheet1", "Sheet2", etc.)

# CRITICAL: File Modification Strategy
- ALWAYS prioritize modifying existing open files IN PLACE rather than creating new files
- The screenshot context shows which file is currently open and should be modified
- For open documents (LibreOffice .docx/.xlsx, text editors, etc.), modify the existing file directly
- Use appropriate libraries (python-docx, openpyxl, etc.) to modify files in place
- CRITICAL: When modifying files, perform COMPLETE OVERWRITES, not appends
- For documents: replace all paragraphs/sheets with new content
- For text files: write the complete new content, overwriting the old
- Only create new files when explicitly required by the task
- Verify your reasoning aligns with the user's intent for the open file

# CRITICAL: Thorough File Inspection Guidelines
- **ALWAYS inspect file contents AND data types before and after modifications**
- Check cell values, formats, data types, number formats, decimal separators, and formatting properties
- For spreadsheets: inspect cell values, number formats, date formats, currency formats, and cell properties
- For documents: inspect text content, formatting, styles, and structural elements
- Verify that modifications actually changed the intended properties (not just values)
- Compare before/after states to ensure changes were applied correctly

# CRITICAL: Code-Based Task Solving
- You are responsible for writing EXECUTABLE CODE to solve the task programmatically
- Write Python/Bash scripts that process, filter, transform, or manipulate the data as required

# CRITICAL: Preserve Document Structure and Formatting
- When modifying documents/spreadsheets, PRESERVE the original structure, headers, and formatting
- NEVER modify column headers, row headers, document titles, or sheet names unless explicitly requested
- Maintain fonts, colors, borders, cell formatting, paragraph styles, etc.
- Only change the content/data, not the structure or visual presentation
- Use libraries that support formatting preservation (python-docx, openpyxl, etc.)
- The goal is to keep the document looking exactly the same, just with different content
- **For column reordering**: Preserve table position - reorder columns within the table without shifting the table itself

# CRITICAL: Final Step Requirement
- At the final step before completing the task (the step before you return DONE), you MUST print out the contents of any files you modified
- Use appropriate commands to display the final state of modified files:
    * For text files: `cat filename` or `head -n 50 filename` for large files
    * For Python files: `cat filename.py`
    * For configuration files: `cat filename.conf`
    * For any other file type: use appropriate viewing commands
- This ensures the user can see exactly what changes were made to the files

# CRITICAL: Verification Instructions
- When you complete a task that modifies files, you MUST provide clear verification instructions
- Include specific details about what the GUI agent should check:
    * Which files were modified and their expected final state
    * What the content should look like (number of lines, key data points, etc.)
    * How to verify the changes are correct
    * Whether the task is complete or if additional GUI actions are needed
- This helps the GUI agent understand what to expect and how to verify your work correctly

# Response Format:
You MUST respond using exactly this format:

<thoughts>
Your step-by-step reasoning about what needs to be done and how to approach the current step.
</thoughts>

<answer>
Return EXACTLY ONE of the following options:

For Python code:
```python
your_python_code_here
```

For Bash commands:
```bash
your_bash_commands_here
```

For task completion:
DONE

For task failure:
FAIL
</answer>

# Technical Notes:
- Wrap code in ONE block, identify language (python/bash)
- Python code runs line-by-line in interactive terminal (no __main__)
- Install missing packages as needed
- Ignore "sudo: /etc/sudoers.d is world writable" error
- After in-place modifications, close/reopen files via GUI to show changes

Focus on progress within your step budget.
\end{lstlisting}
\tcbsubtitle{User Message}
\begin{lstlisting}[language=,basicstyle=\ttfamily\scriptsize,breaklines=true,breakindent=0pt,frame=none,backgroundcolor=\color{green!3!white},xleftmargin=0pt,xrightmargin=0pt,aboveskip=2pt,belowskip=2pt]
Task: {task_instruction}

Current screenshot is provided for context.
\end{lstlisting}
\end{tcolorbox}

\begin{tcolorbox}[enhanced jigsaw, enforce breakable, pad at break=2mm, colback=green!3!white, colframe=green!50!black, left=1mm, right=1mm, top=1mm, bottom=1mm, title=Prompt 2e: Code Agent Summary Prompt]
\small
\tcbsubtitle{System Prompt}
\begin{lstlisting}[language=,basicstyle=\ttfamily\scriptsize,breaklines=true,breakindent=0pt,frame=none,backgroundcolor=\color{green!3!white},xleftmargin=0pt,xrightmargin=0pt,aboveskip=2pt,belowskip=2pt]
You are a code execution summarizer. Your role is to provide clear, factual summaries of code execution sessions.

Key responsibilities:
- Summarize the code logic and approach used at each step
- Describe the outputs and results produced by code execution
- Explain the progression of the solution approach
- Use neutral, objective language without making judgments about success or failure
- Focus on what was attempted and what resulted
- Keep summaries concise and well-structured

CRITICAL: Include verification instructions for the GUI agent
- If files were modified, provide specific verification guidance:
  * What files were changed and their expected final state
  * What the GUI agent should look for when verifying
  * How to verify the changes are correct
  * Whether the task appears complete or if additional GUI actions are needed
- This helps the GUI agent understand what to expect and verify your work properly

Always maintain a factual, non-judgmental tone.
\end{lstlisting}
\tcbsubtitle{User Message}
\begin{lstlisting}[language=,basicstyle=\ttfamily\scriptsize,breaklines=true,breakindent=0pt,frame=none,backgroundcolor=\color{green!3!white},xleftmargin=0pt,xrightmargin=0pt,aboveskip=2pt,belowskip=2pt]
{execution_context}

Please provide a concise summary of the code execution session. Focus on:

1. The code logic implemented at each step
2. The outputs and results produced by each code execution
3. The progression of the solution approach

Do not make judgments about success or failure. Simply describe what was attempted and what resulted.

Keep the summary under 150 words and use clear, factual language.
\end{lstlisting}
\end{tcolorbox}

\textbf{Reflection}

The Reflection module monitors the task trajectory and provides feedback on the agent's progress. It observes the task description and the current trajectory, evaluates whether the task is on track, off-track, or completed, and generates a reflection accordingly. Prompt 2f guides the module on how to assess trajectories and emphasizes that file changes or application restarts may be legitimate results of code agent execution. The output is purely diagnostic and does not suggest specific actions.

\begin{tcolorbox}[enhanced jigsaw, enforce breakable, pad at break=2mm, colback=green!3!white, colframe=green!50!black, left=1mm, right=1mm, top=1mm, bottom=1mm, title=Prompt 2f: Reflection Prompt]
\small
\tcbsubtitle{System Prompt}
\begin{lstlisting}[language=,basicstyle=\ttfamily\scriptsize,breaklines=true,breakindent=0pt,frame=none,backgroundcolor=\color{green!3!white},xleftmargin=0pt,xrightmargin=0pt,aboveskip=2pt,belowskip=2pt]
You are an expert computer use agent designed to reflect on the trajectory of a task and provide feedback on what has happened so far.
You have access to the Task Description and the Current Trajectory of another computer agent. The Current Trajectory is a sequence of a desktop image, chain-of-thought reasoning, and a desktop action for each time step. The last image is the screen's display after the last action.

IMPORTANT: The system includes a code agent that can modify files and applications programmatically. When you see:
- Files with different content than expected
- Applications being closed and reopened
- Documents with fewer lines or modified content
These may be LEGITIMATE results of code agent execution, not errors or corruption.

Your task is to generate a reflection. Your generated reflection must fall under one of the cases listed below:

Case 1. The trajectory is not going according to plan. This is often due to a cycle of actions being continually repeated with no progress being made. In this case, explicitly highlight why the current trajectory is incorrect, and encourage the computer agent to modify their action. However, DO NOT encourage a specific action in particular.
Case 2. The trajectory is going according to plan. In this case, simply tell the agent to continue proceeding as planned. DO NOT encourage a specific action in particular.
Case 3. You believe the current task has been completed. In this case, tell the agent that the task has been successfully completed.

To be successful, you must follow the rules below:
- **Your output MUST be based on one of the case options above**.
- DO NOT suggest any specific future plans or actions. Your only goal is to provide a reflection, not an actual plan or action.
- Any response that falls under Case 1 should explain why the trajectory is not going according to plan. You should especially lookout for cycles of actions that are continually repeated with no progress.
- Any response that falls under Case 2 should be concise, since you just need to affirm the agent to continue with the current trajectory.
- IMPORTANT: Do not assume file modifications or application restarts are errors - they may be legitimate code agent actions
- Consider whether observed changes align with the task requirements before determining if the trajectory is off-track
\end{lstlisting}
\tcbsubtitle{User Message}
\begin{lstlisting}[language=,basicstyle=\ttfamily\scriptsize,breaklines=true,breakindent=0pt,frame=none,backgroundcolor=\color{green!3!white},xleftmargin=0pt,xrightmargin=0pt,aboveskip=2pt,belowskip=2pt]
{worker_history[-1]}
{screenshot}
\end{lstlisting}
\end{tcolorbox}

\subsubsection{OpenClaw}

\paragraph{Note on simplifications.}
Prompt 3 is shown with two simplifications because the original
OpenClaw system prompt is over $33{,}000$ characters and $675$ lines:
(i) the \texttt{<available\_skills>} XML block, which lists 53 skills
in the captured request, is rendered with only two example entries, 
the full skill set is given in Table~\ref{tab:skill_library};
(ii) OpenClaw appends seven user-workspace files
(\texttt{AGENTS.md}, \texttt{SOUL.md}, \texttt{TOOLS.md},
\texttt{IDENTITY.md}, \texttt{USER.md}, \texttt{HEARTBEAT.md},
\texttt{BOOTSTRAP.md}; about 414 lines in total) verbatim at the end of
the system prompt. These are templates auto-generated by OpenClaw on a
fresh install, and we summarize each file in one line rather than
reproducing its full content. In addition, machine-specific values
(file paths, host name, model ID, etc.) are replaced with placeholders
such as \texttt{<host>} and \texttt{<model>} throughout.

\begin{tcolorbox}[enhanced jigsaw, enforce breakable, pad at break=2mm, colback=orange!3!white, colframe=orange!50!black, left=1mm, right=1mm, top=1mm, bottom=1mm, title=Prompt 3: OpenClaw System Prompt]
\small
\tcbsubtitle{System Prompt}
\begin{lstlisting}[language=,basicstyle=\ttfamily\scriptsize,breaklines=true,breakindent=0pt,frame=none,backgroundcolor=\color{orange!3!white},xleftmargin=0pt,xrightmargin=0pt,aboveskip=2pt,belowskip=2pt]
You are a personal assistant running inside OpenClaw.
## Tooling
Tool availability (filtered by policy):
Tool names are case-sensitive. Call tools exactly as listed.
- read: Read file contents
- write: Create or overwrite files
- edit: Make precise edits to files
- exec: Run shell commands (pty available for TTY-required CLIs)
- process: Manage background exec sessions
- web_fetch: Fetch and extract readable content from a URL
- browser: Control web browser
- canvas: Present/eval/snapshot the Canvas
- nodes: List/describe/notify/camera/screen on paired nodes
- cron: Manage cron jobs and wake events (use for reminders; when scheduling a reminder, write the systemEvent text as something that will read like a reminder when it fires, and mention that it is a reminder depending on the time gap between setting and firing; include recent context in reminder text if appropriate)
- message: Send messages and channel actions
- gateway: Restart, apply config, or run updates on the running OpenClaw process
- agents_list: List agent ids allowed for sessions_spawn
- sessions_list: List other sessions (incl. sub-agents) with filters/last
- sessions_history: Fetch history for another session/sub-agent
- sessions_send: Send a message to another session/sub-agent
- subagents: List, steer, or kill sub-agent runs for this requester session
- session_status: Show a /status-equivalent status card (usage + time + Reasoning/Verbose/Elevated); use for model-use questions (session_status); optional per-session model override
- image: Analyze an image with the configured image model
- feishu_app_scopes: List current app permissions (scopes). Use to debug permission issues or check available capabilities.
- feishu_bitable_create_app: Create a new Bitable (multidimensional table) application
- feishu_bitable_create_field: Create a new field (column) in a Bitable table
- feishu_bitable_create_record: Create a new record (row) in a Bitable table
- feishu_bitable_get_meta: Parse a Bitable URL and get app_token, table_id, and table list. Use this first when given a /wiki/ or /base/ URL.
- feishu_bitable_get_record: Get a single record by ID from a Bitable table
- feishu_bitable_list_fields: List all fields (columns) in a Bitable table with their types and properties
- feishu_bitable_list_records: List records (rows) from a Bitable table with pagination support
- feishu_bitable_update_record: Update an existing record (row) in a Bitable table
- feishu_doc: Feishu document operations. Actions: read, write, append, create, list_blocks, get_block, update_block, delete_block
- feishu_drive: Feishu cloud storage operations. Actions: list, info, create_folder, move, delete
- feishu_wiki: Feishu knowledge base operations. Actions: spaces, nodes, get, create, move, rename
- memory_get: Safe snippet read from MEMORY.md or memory/*.md with optional from/lines; use after memory_search to pull only the needed lines and keep context small.
- memory_search: Mandatory recall step: semantically search MEMORY.md + memory/*.md (and optional session transcripts) before answering questions about prior work, decisions, dates, people, preferences, or todos; returns top snippets with path + lines. If response has disabled=true, memory retrieval is unavailable and should be surfaced to the user.
- sessions_spawn: Spawn a sub-agent session
- tts: Convert text to speech. Audio is delivered automatically from the tool result -- reply with NO_REPLY after a successful call to avoid duplicate messages.
- voice_call: Make phone calls and have voice conversations via the voice-call plugin.
TOOLS.md does not control tool availability; it is user guidance for how to use external tools.
For long waits, avoid rapid poll loops: use exec with enough yieldMs or process(action=poll, timeout=<ms>).
If a task is more complex or takes longer, spawn a sub-agent. Completion is push-based: it will auto-announce when done.
Do not poll `subagents list` / `sessions_list` in a loop; only check status on-demand (for intervention, debugging, or when explicitly asked).
## Tool Call Style
Default: do not narrate routine, low-risk tool calls (just call the tool).
Narrate only when it helps: multi-step work, complex/challenging problems, sensitive actions (e.g., deletions), or when the user explicitly asks.
Keep narration brief and value-dense; avoid repeating obvious steps.
Use plain human language for narration unless in a technical context.
## Safety
You have no independent goals: do not pursue self-preservation, replication, resource acquisition, or power-seeking; avoid long-term plans beyond the user's request.
Prioritize safety and human oversight over completion; if instructions conflict, pause and ask; comply with stop/pause/audit requests and never bypass safeguards.
Do not manipulate or persuade anyone to expand access or disable safeguards. Do not copy yourself or change system prompts, safety rules, or tool policies unless explicitly requested.
## OpenClaw CLI Quick Reference
OpenClaw is controlled via subcommands. Do not invent commands.
To manage the Gateway daemon service (start/stop/restart):
- openclaw gateway status
- openclaw gateway start
- openclaw gateway stop
- openclaw gateway restart
If unsure, ask the user to run `openclaw help` (or `openclaw gateway --help`) and paste the output.
## Skills (mandatory)
Before replying: scan <available_skills> <description> entries.
- If exactly one skill clearly applies: read its SKILL.md at <location> with `read`, then follow it.
- If multiple could apply: choose the most specific one, then read/follow it.
- If none clearly apply: do not read any SKILL.md.
Constraints: never read more than one skill up front; only read after selecting.
Remote macOS node available (iMac Pro). Run macOS-only skills via nodes.run on that node.

The following skills provide specialized instructions for specific tasks.
Use the read tool to load a skill's file when the task matches its description.

<available_skills>
  <skill>
    <name>apple-notes</name>
    <description>Manage Apple Notes via the `memo` CLI on macOS (create, view, edit, delete, search, move, and export notes).</description>
    <location>~/Codes/openclaw/skills/apple-notes/SKILL.md</location>
  </skill>
  <skill>
    <name>apple-reminders</name>
    <description>Manage Apple Reminders via remindctl CLI (list, add, edit, complete, delete).</description>
    <location>~/Codes/openclaw/skills/apple-reminders/SKILL.md</location>
  </skill>
  ... <!-- remaining skills omitted; the full list of 53 skills is given in Table 9 -->
</available_skills>

## Memory Recall
Before answering anything about prior work, decisions, dates, people, preferences, or todos: run memory_search on MEMORY.md + memory/*.md; then use memory_get to pull only the needed lines. If low confidence after search, say you checked.
Citations: include Source: <path#line> when it helps the user verify memory snippets.
## OpenClaw Self-Update
Get Updates (self-update) is ONLY allowed when the user explicitly asks for it.
Do not run config.apply or update.run unless the user explicitly requests an update or config change; if it's not explicit, ask first.
Actions: config.get, config.schema, config.apply (validate + write full config, then restart), update.run (update deps or git, then restart).
After restart, OpenClaw pings the last active session automatically.
If you need the current date, time, or day of week, run session_status (session_status).
## Workspace
Your working directory is: /Users/<user>/.openclaw/workspace
Treat this directory as the single global workspace for file operations unless explicitly instructed otherwise.
Reminder: commit your changes in this workspace after edits.
## Documentation
OpenClaw docs: <local-docs-path>
Mirror: https://docs.openclaw.ai
Source: https://github.com/openclaw/openclaw
Community: https://discord.com/invite/clawd
Find new skills: https://clawhub.com
For OpenClaw behavior, commands, config, or architecture: consult local docs first.
When diagnosing issues, run `openclaw status` yourself when possible; only ask the user if you lack access (e.g., sandboxed).
## Current Date & Time
Time zone: America/Los_Angeles
## Workspace Files (injected)
These user-editable files are loaded by OpenClaw and included below in Project Context.
## Reply Tags
To request a native reply/quote on supported surfaces, include one tag in your reply:
- Reply tags must be the very first token in the message (no leading text/newlines): [[reply_to_current]] your reply.
- [[reply_to_current]] replies to the triggering message.
- Prefer [[reply_to_current]]. Use [[reply_to:<id>]] only when an id was explicitly provided (e.g. by the user or a tool).
Whitespace inside the tag is allowed (e.g. [[ reply_to_current ]] / [[ reply_to: 123 ]]).
Tags are stripped before sending; support depends on the current channel config.
## Messaging
- Reply in current session -> automatically routes to the source channel (Signal, Telegram, etc.)
- Cross-session messaging -> use sessions_send(sessionKey, message)
- Sub-agent orchestration -> use subagents(action=list|steer|kill)
- `[System Message] ...` blocks are internal context and are not user-visible by default.
- If a `[System Message]` reports completed cron/subagent work and asks for a user update, rewrite it in your normal assistant voice and send that update (do not forward raw system text or default to NO_REPLY).
- Never use exec/curl for provider messaging; OpenClaw handles all routing internally.
### message tool
- Use `message` for proactive sends + channel actions (polls, reactions, etc.).
- For `action=send`, include `to` and `message`.
- If multiple channels are configured, pass `channel` (telegram|whatsapp|discord|irc|googlechat|slack|signal|imessage|feishu).
- If you use `message` (`action=send`) to deliver your user-visible reply, respond with ONLY: NO_REPLY (avoid duplicate replies).
- Inline buttons not enabled for whatsapp. If you need them, ask to set whatsapp.capabilities.inlineButtons ("dm"|"group"|"all"|"allowlist").

# Project Context
The following project context files have been loaded:
If SOUL.md is present, embody its persona and tone. Avoid stiff, generic replies; follow its guidance unless higher-priority instructions override it.

## AGENTS.md
Workspace conventions: how to read SOUL.md / USER.md / memory files on each session, how to use MEMORY.md as long-term memory, group-chat behavior, when to use emoji reactions, what to write down and when. (~213 lines)

## SOUL.md
Agent persona and core principles: "be genuinely helpful, not performatively helpful", have opinions, be resourceful before asking, earn trust through competence, respect privacy, never send half-baked replies. (~37 lines)

## TOOLS.md
Template for user-specific environment notes (camera names, SSH hosts, preferred TTS voices, speaker/room names). Separated from skills so user configuration does not leak. (~41 lines)

## IDENTITY.md
Blank template (name / creature / vibe / emoji / avatar) the agent fills in during first-run bootstrap. (~24 lines)

## USER.md
Blank template for facts about the human (name, pronouns, timezone, context). Populated over time. (~18 lines)

## HEARTBEAT.md
Periodic self-check list. Empty by default; the agent edits this to track recurring tasks. (~6 lines)

## BOOTSTRAP.md
First-run onboarding script: have a conversation, pick a name/creature/vibe/emoji, write IDENTITY.md and USER.md, optionally connect a messaging channel, then delete this file. (~73 lines)

## Silent Replies
When you have nothing to say, respond with ONLY: NO_REPLY
Rules:
- It must be your ENTIRE message -- nothing else
- Never append it to an actual response (never include "NO_REPLY" in real replies)
- Never wrap it in markdown or code blocks
Wrong: "Here's help... NO_REPLY"
Wrong: "NO_REPLY"
Right: NO_REPLY
## Heartbeats
Heartbeat prompt: Read HEARTBEAT.md if it exists (workspace context). Follow it strictly. Do not infer or repeat old tasks from prior chats. If nothing needs attention, reply HEARTBEAT_OK.
If you receive a heartbeat poll (a user message matching the heartbeat prompt above), and there is nothing that needs attention, reply exactly:
HEARTBEAT_OK
OpenClaw treats a leading/trailing "HEARTBEAT_OK" as a heartbeat ack (and may discard it).
If something needs attention, do NOT include "HEARTBEAT_OK"; reply with the alert text instead.
## Runtime
Runtime: agent=main | host=<host> | repo=<workspace> | os=<os> | node=<node-version> | model=<model> | default_model=<model> | shell=<shell> | channel=<channel> | capabilities=<caps> | thinking=<level>
Reasoning: off (hidden unless on/stream). Toggle /reasoning; /status shows Reasoning when enabled.
\end{lstlisting}

\tcbsubtitle{User Message}
\begin{lstlisting}[language=,basicstyle=\ttfamily\scriptsize,breaklines=true,breakindent=0pt,frame=none,backgroundcolor=\color{orange!3!white},xleftmargin=0pt,xrightmargin=0pt,aboveskip=2pt,belowskip=2pt]
Conversation info (untrusted metadata):
```json
{
  "message_id": "<message-id>",
  "sender_id": "openclaw-macos",
  "sender": "openclaw-macos"
}
```

[<timestamp>] {instruction}
\end{lstlisting}
\end{tcolorbox}

\section{Extended analysis}
\label{app:analysis}

\subsection{Per-category results}
\label{app:per_app}

Table~\ref{tab:per_category} reports Pass@1 broken down by task category
for selected framework-model configurations.

\begin{table*}[!htbp]
  \caption{Pass@1 (\%) broken down by task category.}
  \label{tab:per_category}
  \centering
  \small
  \begin{tabular}{llcccccc}
    \toprule
    Framework & Model & Productivity & System & Internet & Development & Multimedia & Multi-App \\
    \midrule
    \multirow{16}{*}{\hfil /\hfil}
  & Claude Opus 4.6      & 18.3 & 40.9 & 50.0 & 65.9 & 81.0 & 20.7 \\                                                                              
  & GPT-5.4              & 57.1 & 52.3 & 51.9 & 77.3 & 74.1 & 48.6 \\                                                                              
  & Gemini 3.1 Pro       & 17.9 & 34.1 & 31.5 & 59.1 & 83.6 & 13.6 \\                                                                              
  & Qwen3VL-235B-A22B    & 27.7 & 29.5 & 17.6 &  6.8 & 34.5 &  6.4 \\                                                                              
  & Qwen3VL-32B          & 22.3 & 27.3 & 17.6 &  9.1 & 41.4 &  7.9 \\                                                                              
  & Qwen3VL-8B           & 21.0 & 18.2 & 12.0 &  2.3 & 22.4 &  2.1 \\                                                                              
  & OpenCUA-32B          & 30.8 & 15.9 & 15.7 &  9.1 & 16.4 &  7.9 \\                                                                              
  & OpenCUA-7B           & 24.6 & 18.2 & 11.1 &  4.5 & 19.8 &  1.4 \\                                                                              
  & GUI-Owl-1.5-32B      & 24.1 & 27.3 & 12.0 &  4.5 & 20.7 &  7.9 \\                                                                              
  & GUI-Owl-1.5-8B       & 15.6 & 22.7 &  4.6 &  2.3 & 11.2 &  3.6 \\                                                                              
  & UI-TARS-72B-DPO      & 23.2 & 15.9 &  2.8 &  6.8 & 19.0 &  1.4 \\                                                                              
  & UI-TARS-1.5-7B       & 18.3 & 13.6 &  0.9 &  2.3 & 12.9 &  1.4 \\                                                                              
  & ScaleCUA-32B         & 19.2 & 13.6 &  2.8 &  6.8 & 12.9 &  0.7 \\                                                                              
  & ScaleCUA-7B          & 17.0 &  6.8 &  0.0 &  2.3 &  2.6 &  0.0 \\                                                                              
  & InternVL3.5-14B         & 11.6 &  9.1 &  0.0 &  0.0 & 10.3 &  0.7 \\                                                                            
  & InternVL3.5-8B          &  9.4 &  6.8 &  0.0 &  0.0 &  6.9 &  0.0 \\                                                                              
  \midrule                                                                                                                                         
  \multirow{8}{*}{AgentS3}                                                                                                                         
  & Claude Opus 4.6      & 68.8 & 63.6 & 68.5 & 88.6 & 86.2 & 40.7 \\                                                                              
  & GPT-5.4              & 54.9 & 56.8 & 61.1 & 75.0 & 80.2 & 41.4 \\                                                                              
  & Gemini 3.1 Pro       & 53.6 & 52.3 & 51.9 & 72.7 & 80.2 & 30.7 \\                                                                              
  & Qwen3VL-235B-A22B    & 46.9 & 38.6 & 46.3 & 40.9 & 77.6 & 25.7 \\                                                                              
  & Qwen3VL-32B          & 36.6 & 54.5 & 35.2 & 34.1 & 57.8 & 19.3 \\                                                                              
  & Qwen3VL-8B           & 25.9 & 25.0 & 25.9 & 25.0 & 58.6 & 12.1 \\                                                                              
  & InternVL3.5-14B         & 16.1 & 15.9 &  5.6 & 11.4 & 30.2 &  1.4 \\                                                                              
  & InternVL3.5-8B          & 14.3 &  9.1 &  3.7 &  4.5 & 19.0 &  0.7 \\                                                                              
  \midrule                                                                                                                                         
  \multirow{8}{*}{OpenClaw}                                                                                                                        
  & Claude Opus 4.6      & 70.1 & 43.2 & 80.6 & 95.5 & 89.7 & 63.6 \\                                                                            
  & GPT-5.4              & 61.6 & 36.4 & 68.5 & 61.4 & 79.3 & 45.0 \\                                                                              
  & Gemini 3.1 Pro       & 62.9 & 38.6 & 71.3 & 81.8 & 87.1 & 40.0 \\                                                                            
  & Qwen3VL-235B-A22B    & 42.9 & 29.5 & 50.9 & 47.7 & 81.0 & 22.9 \\                                                                              
  & Qwen3VL-32B          & 29.5 & 22.7 & 47.2 & 38.6 & 65.5 &  5.7 \\                                                                              
  & Qwen3VL-8B           & 22.3 & 22.7 & 27.8 & 25.0 & 66.4 &  2.9 \\                                                                              
  & InternVL3.5-14B      &  0.0 &  0.0 &  0.0 &  0.0 &  0.0 &  0.0 \\                                                                              
  & InternVL3.5-8B    &  0.0 &  0.0 &  0.0 &  0.0 &  0.0 &  0.0 \\  
    \bottomrule
  \end{tabular}
\end{table*}

\subsection{Failure mode analysis}
\label{app:failures}

We analyze recurring failure patterns observed across agent frameworks and models, drawing on execution trajectories from baseline, AgentS3, and OpenClaw evaluations.

\paragraph{Text input and focus failures.}
The most pervasive failure across all agents under the baseline framework involves text not reaching the intended UI element. The \texttt{pyautogui.write()} function types characters one at a time, and if the target field loses focus mid-typing, text is partially entered or misdirected. For example, GPT-5.4 on \texttt{numbers/1\_1} produced corrupted table titles like ``Tabatable 111le 1'' when \texttt{Cmd+A} failed to select all existing text before overwriting. Claude Opus~4.6 on the baseline exhibited similar issues in Reminders, spending 10+ steps clicking the same text field coordinates without the field entering edit mode.

\paragraph{Repetitive action loops.}
Weaker models frequently enter degenerate loops, repeating identical actions across dozens of steps. Qwen3VL-32B on \texttt{new\_obsidian/1\_1} issued the same \texttt{find} command 40 consecutive times (Steps 11--50) because it could not determine from screenshots that the terminal window was not focused. Claude Opus~4.6 on the baseline showed a milder variant on \texttt{reminders/1\_1}, clicking the same sidebar coordinates 10+ times while recognizing the action was failing but unable to formulate an alternative strategy.

\paragraph{App navigation and dialog failures.}
Models struggle with macOS-specific UI patterns such as template choosers, multi-level menus, and segmented controls. Claude Opus~4.6 under the baseline on \texttt{keynote/1\_2} double-clicked the wrong position in the template chooser, creating a document with ``Basic White'' instead of ``Basic Black.'' On \texttt{pages/2\_1}, a double-click intended to select the Blank template accidentally navigated to the Newsletters category instead.

\paragraph{Coordinate precision errors.}
Small UI controls like toggle switches and time pickers require pixel-level accuracy that agents frequently miss. GPT-5.4 on \texttt{new\_reminders/2\_1} needed five attempts at nearly identical coordinates (varying by 1--6 pixels) to activate a Time toggle switch. The Clock app's segmented time picker and alarm sound dropdown are particularly challenging for all agents.

\paragraph{Tool-use failures (OpenClaw).}
In the OpenClaw framework, weaker models frequently call nonexistent tools by guessed names. Qwen3VL-8B on \texttt{new\_himalaya/1\_1} attempted to call a ``himalaya'' tool twice, received ``Tool not found'' errors, then wrote a placeholder answer (``otp\_code: none'') without actually checking any emails, a \emph{premature surrender} pattern. On \texttt{keynote/1\_1}, the same model incorrectly concluded the task was impossible via CLI, despite having UI automation tools available. InternVL3.5 models exhibited an extreme version: responding with \texttt{NO\_REPLY} to 67\% of OpenClaw tasks due to inability to parse the tool-use protocol.

\paragraph{Multi-step reasoning failures.}
Complex tasks requiring state tracking across multiple applications expose reasoning limitations. InternVL3.5-14B on \texttt{new\_github/1\_1} typed a \texttt{curl} command into Safari's address bar instead of Terminal, then interpreted the resulting Google search page as though the command had executed. GPT-5.4 on \texttt{new\_reminders/2\_1} embedded a date string in the reminder title rather than setting it as a due date property, creating a malformed reminder that required additional steps to clean up.

\paragraph{Timeout and max-step exhaustion.}
Agents frequently exhaust the 50-step limit during open-ended exploration. GPT-5.4 on \texttt{new\_blogwatcher/5\_1} spent 50 steps searching for an article: 6 steps opening apps, 5 steps port-scanning localhost, 12 steps navigating dashboards, and 27 steps running \texttt{grep} with different keyword patterns, none of which matched the target article title.
  
\subsection{Example task trajectories}
\label{app:trajectories}

\begin{figure*}[!htbp]
  \centering
  \includegraphics[width=\linewidth]{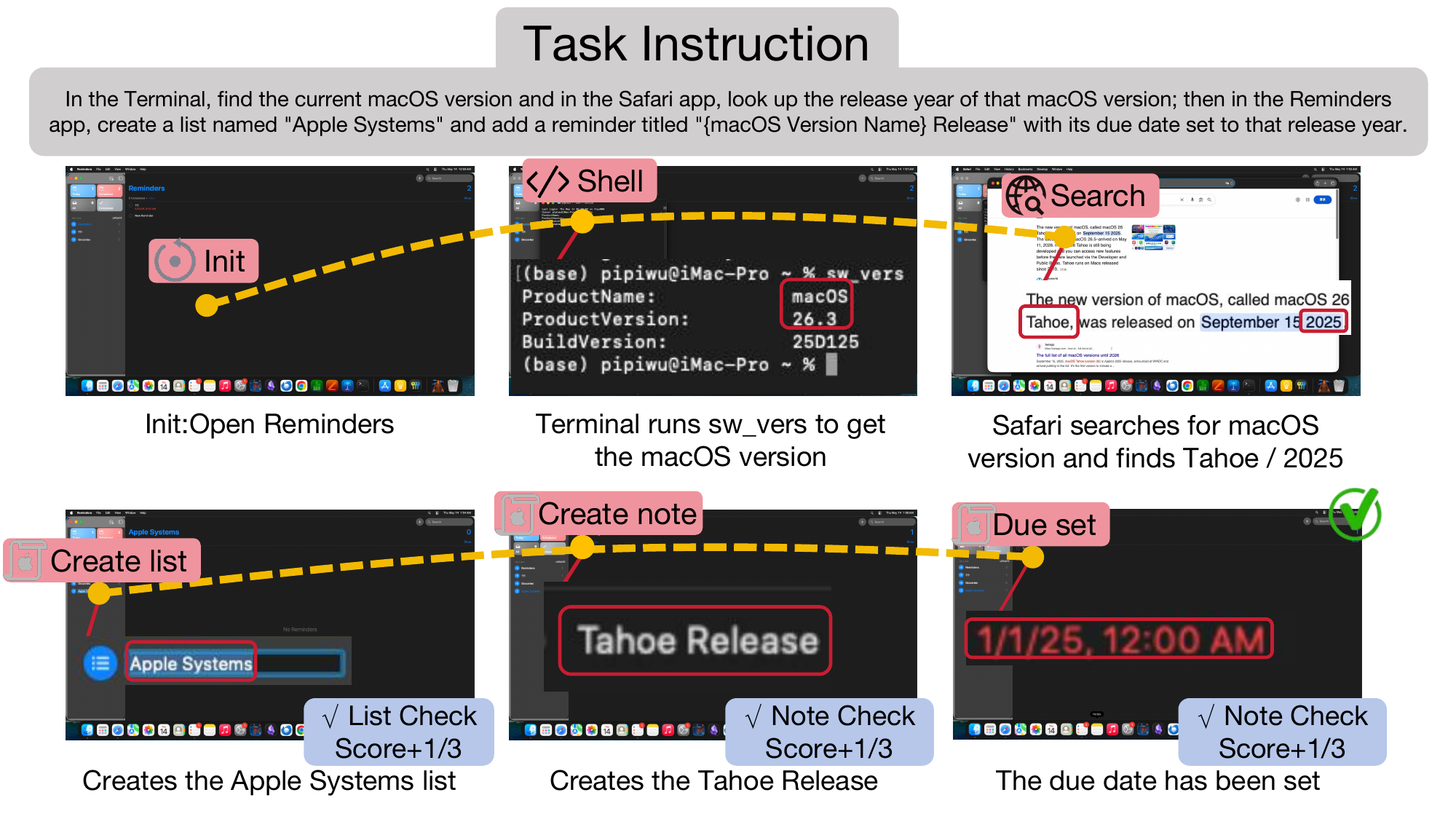}
  \caption{Example task trajectory from GPT-5.4 under the baseline on a
  multi-application task. The agent uses both CLI (\texttt{sw\_vers}
  in Terminal) and GUI (Safari search; Reminders interaction) to
  complete the task. Three fine-grained checkpoints (list creation,
  reminder creation, and due-date assignment) each contribute
  approximately one-third of the score, all satisfied in this run.}
  \label{fig:trajectory_example}
\end{figure*}

We provide a representative trajectory in
Figure~\ref{fig:trajectory_example} to illustrate three properties of
MacAgentBench tasks: long-horizon, cross-application execution; mixed
GUI and CLI interaction; and fine-grained multi-checkpoint scoring.
The task spans Terminal, Safari, and Reminders. Running
GPT-5.4 on the baseline, the agent (i) runs \texttt{sw\_vers} in Terminal
to obtain the current macOS version, (ii) searches in Safari for the
corresponding code name and release year (Tahoe / 2025), and
(iii) creates the \texttt{Apple Systems} list in Reminders, adds a
\texttt{Tahoe Release} reminder, and sets its due date to the release
year. The trajectory is scored against three checkpoints (list creation, reminder creation, and due-date assignment), each contributing
roughly one-third of the score; all three are satisfied
in this run, yielding a final score of $1.0$.

\end{document}